\documentclass[11pt]{article}
\usepackage[margin=1in]{geometry}
\usepackage[round]{natbib}
\usepackage{times}
\usepackage{microtype}
\usepackage[utf8]{inputenc}
\usepackage[T1]{fontenc}
\usepackage{amsmath,amssymb,amsfonts}
\usepackage{graphicx}
\usepackage{booktabs}
\usepackage{tabularx}
\usepackage{array}
\usepackage{hyperref}
\usepackage{url}

\title{Give Up to Get More: Priors, Not Policies, Govern Test-Time Compute Allocation}
\author{Jinhyung Bae\\ Department of Industrial and Management Engineering\\ Hankuk University of Foreign Studies, Yongin, Republic of Korea\\ \texttt{kevinbae2006@hufs.ac.kr}}
\date{}

\begin{document}
\maketitle

\begin{abstract}
When many inputs share a budget of repeated attempts by a large language model (LLM), a common rule gives harder inputs more. On public math, code and formal-proof samples, the best published difficulty-proportional rule captures half or more of the oracle gain on benchmarks the model mostly solves, but at most 13\% on hard workloads, even given true pass rates. Because the value of one more attempt peaks at intermediate difficulty, the optimal static allocation is inverted-U and gives little to inputs that are hopeless at the current budget. In a strict survey of 221 LLM allocation papers, at least 13 give the hardest inputs the most attempts; their easier settings sit at budgets where this is near-optimal, so the cost appears on harder workloads. Acting on it requires knowing which inputs are solvable, so information, not the policy, drives the gain. With a verifier and a prior from a related model's past runs on the same inputs, Thompson Sampling raises the solved share on the hardest workload from 14\% to 27\%; without such a prior, no policy we tested beats round-robin. For the policies we tested, the gain vanishes between prior rank correlations of about 0.3 and 0.55, which cheap text-based priors barely reach on the weakest models, and it requires a verifier whose false acceptances are rare relative to the pass rate. On new Qwen2.5 samples the pattern holds, and tempering the prior avoids losses on easy workloads. We turn these results into a decision guide.
\end{abstract}

\section{Introduction}\label{sec:1}

Training and evaluating reasoning models both spend compute on repeated attempts at a fixed set of inputs, and the budget must be divided across those inputs. Reinforcement learning (RL) with verifiable rewards samples many rollouts per prompt from a fixed pool and rewards those that pass an answer check or unit tests (Lambert et al., 2024; DeepSeek-AI, 2025). Each resulting checkpoint is then evaluated by sampling repeatedly on fixed benchmarks, where either a verifier accepts a candidate or the best one is selected (Brown et al., 2024; Snell et al., 2025). In both cases many inputs share one budget, and someone must decide how many attempts each input receives. A natural answer, adopted by at least 13 of the 221 papers we survey (Section~\ref{sec:3}), is to estimate each input's difficulty and give harder inputs more attempts. We ask when that answer is right.

Giving harder inputs more attempts is close to optimal on workloads a model mostly solves, but it recovers almost none of the attainable gain on workloads the model mostly fails. On the former, the best rule from the literature, a calibration rule (IAS) that gives each input enough attempts to reach a target success probability up to a cap, captures 50--100\% of the gain that an oracle allocation achieves over uniform allocation. On the smallest models on MATH (Pythia) and on code generation, the same rules capture about 0\% of a 5--8 point headroom, although they are given the true pass rates, and tiered variants do no better, and usually worse, than uniform allocation (Section~\ref{sec:4}). What makes a workload hard here is the product of budget and pass rate, not accuracy alone, so a benchmark with modest single-attempt accuracy can still favor ``harder gets more'' at a large budget (Section~\ref{sec:2}, Section~\ref{sec:6}).

The failure follows from a principle that optimal search theory, crowdsourcing and solver scheduling already use: the value of one more attempt peaks at intermediate difficulty. For a binary success objective, the optimal static allocation is therefore inverted-U in difficulty. It gives nothing to inputs that are hopeless at the current budget, rises with solvability, and gives harder inputs more only past a threshold (Section~\ref{sec:2}). ``Harder gets more'' matches only this last branch, which dominates when the budget is large relative to difficulty. Crowdsourcing reached the rule explicitly, putting ``highly ambiguous instances aside to save budget'' (Chen et al., 2013). Related results appear in restart scheduling for SAT solvers (Kautz et al., 2002; Streeter et al., 2007) and in optimal search theory (Koopman, 1957; Stone, 1975). Neyman's (1934) allocation and Atkinson's (1970) inequality index give the analogue for continuous objectives (Section~\ref{sec:2}, Appendix~\ref{app:A}). Across twelve other fields (Section~\ref{sec:3}), areas whose methods build on this classical allocation literature apply these principles, while areas that adopted adaptive compute without that connection, including LLM test-time compute, use ``harder gets more''.

When a verifier confirms success, the decision becomes which unsolved input to attempt next, and the answer depends more on information about the inputs than on the policy. Without input-specific information, round-robin over unsolved inputs is optimal, and Thompson Sampling (TS), an upper confidence bound (UCB) rule and a Bayes-myopic index all reduce to it (Section~\ref{sec:5}). That an informative prior helps is therefore expected; the practical question is how informative it must be. With a prior built from a related model's past pass rates on the same inputs, all three policies gain on the hardest workloads, by up to 14 points. When we add noise to this prior on a fixed workload, the gain falls smoothly with the prior's rank correlation with true pass rates and, for TS, vanishes between 0.3 and 0.4. This gives a break-even level that can be checked on a calibration set before a prior is used. The cheap priors we tried for new inputs, namely text features, LLM-rated difficulty and charged samples from smaller models, lie near or below this level on the hardest workloads and gave at most 1.7 points at 16 attempts per problem. The choice among the three policies matters only at second order (Section~\ref{sec:5}).

The gain also requires a verifier whose false acceptance rate is well below the workload's pass rate: it survives 20\% false rejections and charging attempts by length, but it all but vanishes at a 5\% false acceptance rate, and most learned reward models we tested on hard problems barely separate correct from incorrect answers.

These conditions match how reasoning models are trained and evaluated, because both repeatedly run the same inputs against a programmatic verifier. In RL with verifiable rewards, the prompt pool is revisited in every epoch, rollouts are sampled independently, and rewards come from answer checks or unit tests, so the previous epoch's pass rates are a history prior at no extra cost. Evaluation suites are re-run for every checkpoint, prompt template or release, and regression tests compare model versions on fixed internal sets. Rollout allocation for RL is also where the principle is closest to being known: methods there already drop prompts with pass rates of 0 or 1 or favor intermediate ones (Yu et al., 2025; Bae et al., 2025; Section~\ref{sec:3}), whereas inference-time allocation rarely does. In these settings the sampling cost of the history has already been paid by runs that happen anyway. Storing per-input pass rates, matching them to the inputs of a new run and calibrating them to the target model and verifier still require engineering effort, which we do not measure. For genuinely novel inputs, none of the cheap priors we evaluated helped, and the recommended policy then defaults to round-robin, which costs nothing but also gains nothing.

The paper is an applied study for anyone who divides a repeated-attempt budget across a fixed set of inputs. It proposes no new learning method; instead, it answers three practical questions: when reallocating attempts can pay, which allocation rule to use, and what information about the inputs is needed. It makes four contributions.
\begin{enumerate}
\item A measurement, on public stored samples, of when difficulty-proportional allocation fails: the best published rule captures half or more of the oracle gain on workloads the model mostly solves and at most 13\% on hard ones, even given true pass rates. The workload's mean pass rate at the planned budget predicts the failure on our workloads; the left-branch share derived in Section~\ref{sec:2} serves as a diagnostic that tracks it about as well.
\item A controlled measurement of how much prior information a verifier-based policy needs. Degrading a history prior on a fixed workload shows that the gain over round-robin falls smoothly with the prior's rank correlation and gives a break-even level to check before use. With an undegraded history prior, any of three standard policies nearly doubles the solved share on the hardest workloads at 16 attempts per problem, and a tempered prior avoids losses on easy workloads. We also report negative results for cheap priors and give a model that predicts when verifier errors erase the gain.
\item A cross-field map and a strict survey of 221 LLM allocation papers, using full-text re-judgment, partial double-coding and an author-reviewed, LLM-assisted check of every candidate, that locate where the principle is known and where it has not yet been connected.
\item A decision guide that turns these results into checks a practitioner can run before choosing a rule (Section~\ref{sec:6}).
\end{enumerate}

We claim no priority for the allocation principle, which is classical and credited where used. We do not show that the surveyed methods are wrong: on easy benchmarks ``harder gets more'' is close to optimal, and the cost we measure appears only when such rules are applied to hard workloads. Nor do we claim that any one policy is best; we recommend Thompson Sampling because its tempered version is the one we tested. Our evidence comes mainly from offline replays of public stored samples on tasks with binary, verifiable success, with one replication on freshly generated samples from recent Qwen2.5 models. The tasks are math, code and formal proofs, and the models are open models from 0.5B to 70B parameters, together with stored GPT-4 (PRM800K) and GPT-4o-mini samples; smaller models also serve as charged proxy samplers in the prior comparison (Table~\ref{tab:3}). Our conclusions therefore apply to recurring inputs with programmatic verifiers, as in RL with verifiable rewards, benchmark evaluation and regression testing on math and code. We make no claim about novel user queries, learned verifiers or open-ended generation, and whether the same magnitudes hold for frontier reasoning models remains open.

Section 2 derives the optimal static allocation and the role of a verifier. Section 3 maps where the principle is known. Section 4 measures static rules without a verifier, and Section 5 studies sequential allocation with a verifier, including prior quality and verifier errors. Section 6 collects the practical recommendations and limitations, Section 7 reviews related work, and Section 8 concludes.

\section{Why the optimal allocation is not monotone in difficulty}\label{sec:2}

When attempts are assigned in advance and success is binary, the optimal allocation gives the most attempts to inputs of intermediate difficulty, not to the hardest ones. In the setting of Section 4, input i has pass rate $p_i$, and $k_i$ independent attempts solve it with probability $1-(1-p_i)^{k_i}$. We maximize the expected number of solved inputs subject to $\sum_i k_i = S = uN$, where u is the average number of attempts per input. An attempt need not be a single sample. It can be any self-contained procedure run on the input, such as a chain of self-correction, a tree search or a long reasoning trace, provided that repeated runs on the same input are independent and each succeeds with the same probability $p_i$. The model then describes how to divide whole runs across inputs, which is how pass@k is measured for reasoning models; it does not describe how compute should be spent inside a run. The objective is separable and concave, so greedy allocation on marginal values is optimal. The marginal value of the (k+1)-th attempt is $\Delta(p,k)=p(1-p)^k$, which increases in p up to $p=1/(k+1)$ and decreases afterwards. Writing $r=-\ln(1-p)$, treating k as continuous and requiring k $\ge$ 0, the KKT conditions r$\cdot$e\^{}{$-$rk} = $\lambda$ give

\[ k^\ast(r) = \max\Bigl(0,\ \frac{\ln(r/\lambda)}{r}\Bigr), \]

where $\lambda$ is fixed by the budget (Appendix~\ref{app:C}). Both the greedy argument and this logarithmic form are standard; the solution is closely related to the classical optimal allocation of search effort (Koopman, 1957; Stone, 1975), and we claim no novelty for it.

\begin{figure}[t]
\centering
\includegraphics[width=0.85\linewidth]{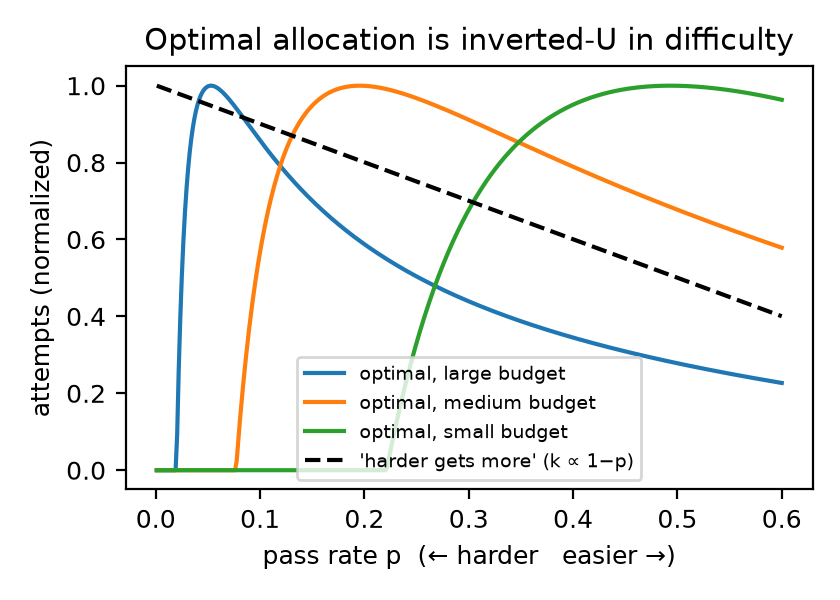}
\caption{Optimal attempts versus pass rate for three budgets, compared with a ``harder gets more'' rule ($k\propto 1-p$). The optimum gives nothing to inputs that are hopeless at the given budget, and its peak moves toward harder inputs only as the budget grows.}\label{fig:1}
\end{figure}

Under known pass rates, independent attempts, a static budget and no verifier, the optimum is therefore inverted-U in difficulty (Figure~\ref{fig:1}). Since r increases with p, thresholds on r are thresholds on the pass rate. Inputs with $r\le\lambda$, that is $p\le 1-e^{-\lambda}$, receive no attempts; allocation rises with solvability up to $r=e\lambda$; and beyond that point allocation falls as p rises, so only among these easier inputs do harder ones receive more. ``Harder gets more'' matches the optimum only on this right branch, which dominates when the budget is large relative to difficulty ($u\cdot p$ large), as on easy benchmarks. Our experiments also give every input at least one attempt. With the constraint k $\ge$ 1, the same conditions give k*(r) = max(1, ln(r/$\lambda$)/r) with $\lambda$ refitted to the budget, so an input receives attempts beyond the first only if r$\cdot$e\^{}{$-$r} > $\lambda$. Both hopeless inputs and nearly certain ones then stop at the minimum, and the peak stays at $r=e\lambda$ (Appendix~\ref{app:C}). The diagnostic below uses the unconstrained form.

On our workloads, the share of inputs on the left branch tracks how much a monotone rule loses. To compute it, one finds $\lambda$ at the planned budget and counts the share of inputs with $\hat r_i < e\lambda$, where $\hat{r}_i$ = $-$ln(1 $-$ $\hat{p}_i$) uses the estimated pass rate $\hat{p}_i$. These are the inputs that the optimum gives nothing or gives fewer attempts than the peak, the left branch of Figure~\ref{fig:1}. Across eleven workload--model combinations, which are not independent because several share a benchmark, this share tracks how much of the headroom a monotone rule captures (Spearman $-$0.92, bootstrap 95\% CI [$-$0.97, $-$0.58]; Pearson $-$0.96; leave-one-out range $-$0.95 to $-$0.89; Appendix~\ref{app:E}). The workload's mean pass rate predicts the captured share at least as well on these workloads (Spearman +0.98). On these workloads the left-branch share has no empirical advantage over the mean pass rate. We report it because it follows from the optimum and, unlike the mean, responds to skewed or bimodal difficulty, but we have not shown that this matters in practice. With eleven related combinations, this is evidence of association on these workloads, not a validated predictor for new ones.

A verifier changes the decision problem from how many attempts to assign in advance to which unsolved input to attempt next. This is the setting of Section~\ref{sec:5}. If success can be checked, attempts are made one at a time and solved inputs leave the pool. Each failure lowers an input's posterior success probability, so each input is a deteriorating arm (an option whose value can only fall as it is used), and attempting next where this probability is highest is a classical index policy (Gittins, 1979; Weitzman, 1979; see the survey of Beyhaghi \& Cai, 2023). We assume that attempts are independent and cost the same, and that the verifier is effectively perfect and free. Under these assumptions and an exchangeable prior, all unsolved inputs with the same number of failures share one index, so the policy is round-robin over unsolved inputs (RR). Any improvement over RR must come from information that distinguishes the inputs.

All results in this paper concern binary success. For continuous objectives such as a reward-model score, the analogous result follows from classical allocation theory (Neyman, 1934): under the assumptions of Appendix~\ref{app:A}, the gain from reallocation equals an Atkinson (1970) inequality index of difficulty, with Neyman allocation as a special case. Appendix~\ref{app:A} gives the derivation and checks it on combinatorial optimization, hyperparameter optimization and LLM reward-model workloads.

\section{Where the principle is known and where it is not yet connected}\label{sec:3}

Published LLM allocation methods repeatedly use the rule that Section~\ref{sec:4} shows to fail on hard workloads, whereas fields whose methods build on classical allocation theory do not. This section documents both patterns; it supports Sections~\ref{sec:4}--\ref{sec:5} and is not the main empirical contribution.

We coded 221 LLM papers that allocate inference compute non-uniformly across inputs. Forward citations of ten seed papers on adaptive test-time compute, together with one keyword search, gave 3,016 unique records. A recall-oriented filter kept 1,269, and automated inclusion screening included 221. Each paper was coded with a fixed codebook that records the allocated unit, the objective, the difficulty signal, the direction of allocation, whether decision and evaluation data are separated, and whether the signal's cost is charged. We call a paper harder-gets-more (HGM) if it allocates repeated independent attempts, uses a difficulty-level signal, gives harder inputs more attempts, and applies no discount to hopeless inputs, that is, no rule that reduces or stops attempts on inputs judged unlikely to be solved (for example, early termination). A cap that every input can reach does not count as a discount.

Most of the 221 papers allocate something other than repeated attempts, such as reasoning length, early exit or model routing, so they cannot exhibit this pattern. By first-pass coding, which we did not verify, 98 allocate repeated attempts, and of the 46 among them that state a direction, 32 (about 70\%) give harder inputs more and 14 allocate non-monotonically; the remaining 52 state no direction and are treated as not HGM, which is conservative for a lower bound. The strict definition narrows this further. Thirteen of the 221 papers (5.9\%) meet this strict definition after an LLM-assisted, author-reviewed check of every candidate against its full text, and four more are borderline (17, or 7.7\%). Because inclusion screening was automated and not independently verified, and the full-text checks were LLM-assisted, we treat 13 as an approximate lower bound and the percentages as an indication that the pattern recurs, not as a precise prevalence rate. Twelve of the 13 allocate at inference time, each with its own rule, and one allocates RL training rollouts (Appendix~\ref{app:D}).

The confirmed rules differ in form but share the same direction. One paper samples one response at a time until one meets a confidence threshold, so low-confidence inputs consume more samples (Huang et al., 2025). Another sets $N=\lceil\log(1-C)/\log(1-p)\rceil$, capped at $N_{\max}$, so that low-p inputs reach $N_{\max}$, and its authors note that they ``assign the maximum budget to the hardest problems'' (Park et al., 2025). Others use tiered trace counts of 21/30/45 (Ozsoy, 2026) or sample counts that grow with predicted entropy (Yeom et al., 2026).

The count of 13 is the end of a three-step funnel (Figure~\ref{fig:5} in Appendix~\ref{app:D}). The first pass had flagged 33 candidates, and an automated full-text re-judgment kept 19. Of the 14 it rejected, 12 allocate refinement rounds, reasoning length or model routing rather than repeated attempts. The full-text check then rejected two of the 19: one terminates unpromising queries early, which is a discount, and the other searches over actions at each step rather than repeating attempts. Of the remaining 17, 13 meet the strict definition and four are borderline. In the four borderline papers, the extra compute mixes repeated attempts with verifier routing, model routing or tree branching, or no fixed-budget comparison is reported.

Two checks bound the reliability of the coding, although neither is fully independent of LLM judgment. A blind second coder recoded a stratified random sample of 50 included papers (15 first-pass HGM candidates and 35 others); the two codings agreed on the HGM flag for 41 of the 50 (Cohen's $\kappa$ = 0.51). The second coder flagged 8 as HGM; 7 of these were also first-pass candidates, and the strict re-judgment gives the same verdict for 6 of the 7. The full-text check of the 19 was LLM-assisted: an LLM read each paper and proposed a verdict with section-level evidence, we reviewed every verdict, and we judged the ambiguous cases ourselves. Every disagreement it produced concerned what counts as a repeated attempt or whether early termination is a discount. The same procedure, applied to a random sample of 10 included papers, marked two refinement-round methods as HGM under a broader definition, which our strict definition excludes.

To see where the principle is already used, we also mapped twelve other fields that divide repeated attempts or measurements across items. For each field, LLM agents gathered 20--36 papers, coded them with a shared codebook (320 coded, 251 included) and summarized the field's classical allocation results, marking which claims they had checked against the abstract or full text; we checked every source quoted below. These maps were not built by systematic search, so they locate the principle but do not estimate its prevalence. The dividing line is not the age of a field but whether its allocation methods build on classical allocation theory. Where they do, methods allocate by marginal value and set hopeless items aside; where adaptive compute was adopted without that connection, methods give more to harder inputs. In the first group, survey sampling and A/B testing use Neyman allocation, and simulation optimization uses OCBA (Chen et al., 2000) and the knowledge gradient (Frazier et al., 2008). SAT solving and algorithm portfolios use restarts (Luby et al., 1993) and pruning of long runs (Kautz et al., 2002), and Streeter et al. (2007) greedily maximize ``the number of instances \ldots{} solved per unit time''. Crowdsourcing has Opt-KG (Chen et al., 2013) and Khetan \& Oh (2016). RL rollout allocation is a young area that reaches the same rule by another route: it drops prompts that are always or never solved and favors intermediate pass rates (Yu et al., 2025; Bae et al., 2025), because a group of rollouts that all fail or all succeed gives no policy-gradient signal. The second group includes LLM test-time compute; image and video generation, where none of the inference-time papers among the 14 we coded uses dispersion or marginal gain; and test-time compute for vision-language-action policies. In multi-target directed fuzzing, LeoFuzz assigns more energy to a target that ``is more difficult to reach'' (Liang et al., 2024). In molecular free-energy estimation, one method allocates samples in proportion to the variance of each estimate (Lee et al., 2025), which gives more to the noisiest, hardest-to-estimate states than Neyman allocation would. Neural combinatorial optimization rarely allocates across instances; the closest work selects among solvers per instance by selector confidence (Gao et al., 2025).

In these maps, ``harder gets more'' appears where a community adopted adaptive compute without connecting it to the classical allocation literature, and RL rollout allocation shows that a direct training signal can lead to the right rule without that connection. We document this co-occurrence but do not claim to have established its cause.

\section{Static allocation: the cost of spending on the hopeless}\label{sec:4}

Without a verifier, harder-gets-more rules capture almost none of the attainable gain on hard workloads, even when given the true pass rates; only the calibration rule recovers most of it, and only on easy workloads.

\begin{quote}
\textbf{Setting: no verifier.} The number of attempts per input is fixed before sampling begins and cannot stop on success, as in final-answer math without a checker or in majority voting. Section~\ref{sec:5} turns to the case with a verifier.
\end{quote}

We replay public stored samples, so that every rule is evaluated on identical attempts. The Large Language Monkeys release (Brown et al., 2024) provides 10,000 samples per problem for MATH, GSM8K, CodeContests and MiniF2F across Pythia, Gemma and Llama-3 models, and PRM800K (Lightman et al., 2024) provides about 1,800 samples per problem for 495 MATH problems. The GSM8K GPT-4o-mini samples come from the Weaver release (Saad-Falcon et al., 2025). Each problem's pool pass rate defines its coverage curve, and attempts are i.i.d. draws from the pool. This matches repeated sampling: independent samples from one prompt at a fixed temperature are i.i.d. draws from the model's output distribution, and a pool of thousands approximates that distribution. Our data contain single-sample attempts. Self-correction or search inside an attempt fits the model when the whole procedure is treated as one attempt (Section~\ref{sec:2}), but attempts that condition on earlier attempts at the same input, such as retrying with feedback from a failed run, are outside it and are not tested here.

Every allocation spends u$\cdot$N attempts in total, and every input receives at least one. Coverage is the expected share of problems solved at least once, the mean over inputs of $1-(1-p_i)^{k_i}$, and the baseline is uniform allocation ($k_i$ = u). The oracle maximizes coverage given the true pass rates by assigning attempts greedily on marginal gains, with at most 16u attempts per input; applied to estimated pass rates, the same greedy procedure is the marginal-value rule. We test four rules from the survey, each given the same true pass rates as the oracle and scaled to the same total budget: the calibration rule $N=\lceil\log(1-C)/\log(1-p)\rceil$ with a cap (IAS; C = 0.95, cap set as large as the budget allows), $k\propto 1-p$, $k\propto\exp(-p)$, and 21:30:45 tiers over pass-rate tertiles. Table~\ref{tab:1} reports each rule's captured share, (rule $-$ uniform)/(oracle $-$ uniform). The $k\propto\exp(-p)$ rule, not shown, captures 20--47\% on the five easy workloads and $-$2\% to +6\% on the hard ones.

\begin{table}[t]
\centering\small
\caption{Share of the oracle's coverage gain over uniform allocation, at $u=16$ attempts per problem.}\label{tab:1}
\resizebox{\linewidth}{!}{%
\begin{tabular}{lllllll}
\toprule
workload & N problems & mean p & oracle gain (pts) & IAS & $k\propto 1-p$ & tiers \\
\midrule
MATH, Pythia-1.4B & 128 & 0.013 & +8.4 & 0\% & $-$1\% & $-$34\% \\
MATH, Pythia-6.9B & 128 & 0.019 & +8.1 & 0\% & $-$2\% & $-$40\% \\
CodeContests, Llama-3-8B & 140 & 0.021 & +4.8 & 1\% & $-$1\% & 0\% \\
CodeContests, Llama-3-70B & 140 & 0.074 & +6.0 & 2\% & 2\% & $-$16\% \\
MATH, Gemma-2B & 128 & 0.093 & +7.4 & 13\% & 6\% & $-$34\% \\
MATH, Llama-3-8B & 128 & 0.268 & +6.3 & 50\% & 40\% & 15\% \\
MATH, Llama-3-70B & 128 & 0.479 & +7.6 & 74\% & 61\% & 32\% \\
PRM800K MATH & 495 & 0.486 & +6.1 & 66\% & 57\% & 33\% \\
GSM8K, Llama-3-8B & 127 & 0.768 & +1.9 & 93\% & 71\% & 28\% \\
GSM8K, GPT-4o-mini (Weaver) & 127 & 0.931 & +0.2 & 102\% & 78\% & 12\% \\
\bottomrule
\end{tabular}}
\end{table}

We omit CodeContests with Gemma-2B (mean p < 0.001), where almost no problem is ever solved and every rule captures 0\% of a +1.8-point headroom. On Weaver GSM8K the headroom is only +0.2 points, so IAS's measured share of 102\% reflects a difference from the oracle of under 0.01 points; the oracle is a greedy integer allocation with a cap of 16u attempts per input, so a heuristic can match or marginally exceed it.

The failure lies in the mapping from difficulty to attempts, not in the difficulty estimate. Every rule receives the true pass rates, so this comparison is the most favorable possible to the heuristics. The captured share rises with the workload's pass rate (Table~\ref{tab:1}; Figure~\ref{fig:2} in Appendix~\ref{app:E}), though not strictly monotonically. On hard workloads the optimum concentrates attempts on the relatively solvable inputs, whereas $k\propto 1-p$ spreads them almost evenly, because it is nearly uniform when all p are small, and the tiered rule pushes them toward the hopeless tail.

With realistic, imperfect signals, only the marginal-value rule gains, and only when the signal is informative. We use a similar model's pass rate on the same problems, calibrated by 2-fold cross-fitting (Spearman $\rho$ = 0.66--0.93 with the target's pass rates). Using these estimated pass rates, the marginal-value rule gains +3.3 to +6.6 points on MATH for Pythia models (history from another Pythia model or from Gemma-2B) and +2.6 points on CodeContests for Llama-3-8B, but about zero on CodeContests for Llama-3-70B ($-$1.1 to +0.1). IAS and $k\propto 1-p$ recover about zero, and tiers lose up to 3.1 points. With the coarse official MATH difficulty level as the signal, every rule ties at about +0.7 points.

\section{With a verifier: the prior matters more than the policy}\label{sec:5}

With a verifier, each of the three index policies we tested beats round-robin on the hardest workloads when it has a history prior, and none does without one; the size of the gain follows the quality of the prior.

\begin{quote}
\textbf{Setting: a verifier is available.} Unit tests, a proof checker or an answer key confirm success, so attempts are made one at a time and an input leaves the pool once solved. Because budget is never spent on solved inputs, the baseline to beat is round-robin over unsolved inputs (RR), not uniform allocation.
\end{quote}

We compare three standard policies with RR, each under two priors. The policies are Thompson Sampling (TS; Thompson, 1933), an upper confidence bound (UCB; Auer et al., 2002) on the posterior pass rate, and a Bayes-myopic index. The exchangeable prior is the empirical distribution of pass rates, the same for every input. The history prior uses a related model's pass rate on each input, which we call the history signal: for a given input, the prior is the empirical distribution of pass rates among calibration inputs whose history signal falls in the same quintile. Calibration is cross-fitted, so no test outcome is used. A tempered prior mixes an input's history-bin distribution (weight w) with the exchangeable distribution (weight 1 $-$ w); w = 1 is the history prior, and w = 0 reduces to RR. We also run the surveyed rule online, sampling until success up to a cap that grows with difficulty.

\begin{table}[t]
\centering\small
\caption{Change in solved share relative to RR (points), over 20 replays. The surveyed-rule column comes from a separate run with the same protocol and is measured against that run's own RR, which differs by at most 2.3 points. Tables~\ref{tab:2}, \ref{tab:2b} and \ref{tab:4} use independent replay sets (20, 100 and 100 replays), so the same cell can differ slightly across tables. The full table covers 6 workloads $\times$ 3 budgets (Appendix~\ref{app:E}).}\label{tab:2}
\begin{tabular}{>{\raggedright\arraybackslash}p{0.314\linewidth}>{\raggedright\arraybackslash}p{0.065\linewidth}>{\raggedright\arraybackslash}p{0.065\linewidth}>{\raggedright\arraybackslash}p{0.065\linewidth}>{\raggedright\arraybackslash}p{0.097\linewidth}>{\raggedright\arraybackslash}p{0.065\linewidth}>{\raggedright\arraybackslash}p{0.065\linewidth}>{\raggedright\arraybackslash}p{0.065\linewidth}}
\toprule
workload (history from) & u & RR solved & TS, exch. & \textbf{TS, history} & UCB, hist. & myopic, hist. & surveyed rule \\
\midrule
MATH, Pythia-1.4B (Pythia-6.9B) & 16 & 14.1\% & $-$0.2 & \textbf{+12.5} & +12.7 & +12.1 & 0.0 \\
MATH, Pythia-6.9B (Pythia-1.4B) & 16 & 18.6\% & +0.3 & \textbf{+13.8} & +14.1 & +13.6 & 0.0 \\
MATH, Gemma-2B (Llama-3-8B) & 4 & 23.4\% & +0.1 & \textbf{+6.3} & +6.2 & +5.5 & 0.0 \\
MATH, Gemma-2B (Llama-3-8B) & 64 & 72.4\% & $-$0.1 & \textbf{+2.1} & +0.7 & $-$2.7 & $-$1.6 \\
CodeContests, Llama-3-8B (70B) & 16 & 5.5\% & $-$0.1 & \textbf{+4.0} & +3.8 & +3.7 & 0.0 \\
MiniF2F (Lean), Llama-3-8B (70B) & 16 & 40.0\% & +0.2 & \textbf{+2.1} & +1.7 & +0.2 & $-$4.4 \\
MATH, Llama-3-8B (70B) & 64 & 90.3\% & 0.0 & \textbf{$-$0.6} & $-$1.0 & $-$2.0 & $-$0.4 \\
\bottomrule
\end{tabular}
\end{table}

The confidence intervals separate the two priors cleanly (Table~\ref{tab:2b}, Appendix~\ref{app:E}, 100 replays at $u=16$). Every interval for the exchangeable prior contains or touches zero, and every interval for TS with the history prior excludes zero. On the two moderately hard workloads, MiniF2F and MATH with Llama-3-8B, the myopic index's intervals include zero while those of TS do not.

\begin{figure}[t]
\centering
\includegraphics[width=0.85\linewidth]{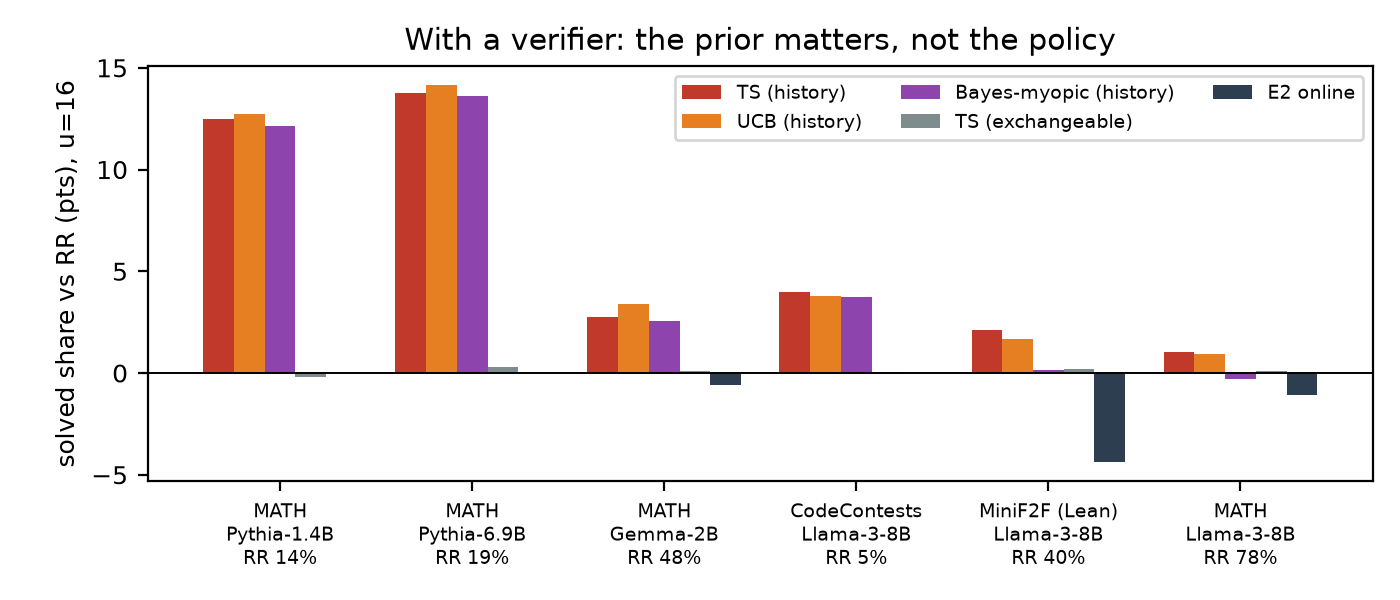}
\caption{Solved share relative to round-robin at $u=16$. With a history prior, all three policies gain on hard workloads. Without it (grey), none does. The surveyed rule (dark) never helps.}\label{fig:3}
\end{figure}

Across all eighteen workload $\times$ budget cells, policies gain only with the history prior, and the differences among them are small. First, without the history prior no policy beats RR by more than 0.3 points: UCB reproduces RR exactly, and TS stays within $-$0.5 to +0.3 (full table in Appendix~\ref{app:E}). Second, with the history prior all three policies gain on hard workloads, by up to 14.1 points (13.8 for TS), which nearly doubles the solved share of Pythia models on MATH at $u=16$; the 100-replay comparison with bootstrap confidence intervals (Appendix~\ref{app:E}) gives up to 14.2 (TS 13.7). Third, on easy workloads with large budgets the myopic index lost most (worst cell $-$2.7 points, against $-$1.0 for UCB and $-$0.6 for TS). The surveyed rule never beats RR and loses 4.4 points on formal proofs.

The prior therefore decides whether there is anything to gain, and the policy matters only at second order. At $u=16$, every policy gains 12--14 points on the hardest workloads with a good prior (0.2--4.0 points on CodeContests and MiniF2F; TS 2.1--4.0) and about zero without one, whereas on stored samples TS and UCB differ by at most 2 points in any cell (Appendix~\ref{app:E}) and only the myopic index falls further behind (by 4.8 points on Gemma-2B at $u=64$). The exception is an easy workload with an imperfect prior, where policy choice and tempering matter (Tables~\ref{tab:9} and \ref{tab:10}). ``Priors, not policies'' summarizes this difference in magnitude; it does not mean that the policy is irrelevant.

Cheap priors fail because their rank correlation with true pass rates is too low (Table~\ref{tab:3}). Text regressions, LLM-derived features and a dissimilar model's history all correlate weakly with true pass rates and give no gain, and fresh proxy samples lose because their cost is charged. Only the past runs of a related model correlate strongly enough to help.

\begin{table}[t]
\centering\small
\caption{Priors tried on the hard workloads (mean pass rate below 0.1) at $u=16$, except the dissimilar-model row, which uses MATH with Llama-3-8B (mean pass rate 0.27). The related-model gains are from Table~\ref{tab:2b}.}\label{tab:3}
\begin{tabular}{>{\raggedright\arraybackslash}p{0.605\linewidth}>{\raggedright\arraybackslash}p{0.173\linewidth}>{\raggedright\arraybackslash}p{0.147\linewidth}}
\toprule
prior source & Spearman with true p & gain vs RR \\
\midrule
TF-IDF regression on problem text & 0.05--0.23 & $\le$ 0 \\
LLM features (bge-large embedding; Qwen2.5-1.5B difficulty rating, P(solvable), NLL) & 0.10--0.46 & $-$2.1 to $-$0.1 ($\approx$ 0 when tempered) \\
Zero-shot difficulty judgments by strong LLMs (Gemini, Claude; ratings or listwise rankings of the problem text) & 0.36--0.40 (Pythia), 0.58 (Gemma-2B) & $-$0.1 to +1.7 \\
Past runs of a dissimilar model (Pythia history for Llama-3-8B) & 0.42 & $-$1.1 \\
16 fresh samples from a smaller model, cost charged & --- & $-$1.7 to $-$4.2 \\
\textbf{Past runs of a related model} & \textbf{0.69--0.93} & \textbf{+3.4 to +13.7} \\
\bottomrule
\end{tabular}
\end{table}

Difficulty judged from problem text by a strong LLM tracks what strong models find hard, not what weak models fail. We asked Gemini and Claude to rate or rank the 128 MATH problems, without access to any model's outcomes. The two raters agree closely (Spearman 0.89), and their judgments correlate with the pass rates of Llama-3-70B at 0.77--0.81, of Llama-3-8B at 0.69--0.71, of Gemma-2B at 0.58--0.60 and of the Pythia models at only 0.36--0.40. Showing the rater twenty problems that Pythia-1.4B solved or failed raised the correlation only from 0.37 to 0.40. Used as a prior, these judgments gave at most +1.7 points on the Pythia workloads at $u=16$ and +2.7 on Gemma-2B at $u=4$, far below the related-model history. The cheapest substitute for history therefore fails exactly where the gain is largest: on the weakest models, whose failures are not visible in the problem text.

\begin{table}[t]
\centering\small
\caption{Controlled degradation: logit noise added to the history signal of a fixed workload, then the prior is rebuilt ($u=16$, 100 replays). Policy columns give the mean gain over RR; the last two columns give paired differences with bootstrap 95\% CIs.}\label{tab:4}
\resizebox{\linewidth}{!}{%
\begin{tabular}{lllllll}
\toprule
workload & rank corr. of prior & TS & UCB & myopic & TS $-$ UCB & TS $-$ myopic \\
\midrule
MATH, Pythia-1.4B & 0.93 & +12.2 & +12.6 & +12.5 & $-$0.4 [$-$0.6, $-$0.1] & $-$0.3 [$-$0.7, +0.1] \\
 & 0.83 & +8.5 & +8.8 & +8.7 & $-$0.3 [$-$0.7, $-$0.0] & $-$0.2 [$-$0.6, +0.2] \\
 & 0.67 & +4.2 & +4.6 & +5.1 & $-$0.4 [$-$0.8, $-$0.1] & $-$0.8 [$-$1.3, $-$0.3] \\
 & 0.50 & +1.9 & +2.4 & +2.5 & $-$0.6 [$-$0.9, $-$0.3] & $-$0.6 [$-$1.1, $-$0.2] \\
 & 0.42 & +0.6 & +1.1 & +1.4 & $-$0.5 [$-$0.8, $-$0.2] & $-$0.7 [$-$1.1, $-$0.4] \\
 & 0.29 & $-$0.8 & $-$0.2 & $-$0.2 & $-$0.6 [$-$0.9, $-$0.2] & $-$0.6 [$-$1.1, $-$0.1] \\
 & 0.19 & $-$1.1 & $-$0.9 & $-$0.9 & $-$0.2 [$-$0.5, +0.2] & $-$0.2 [$-$0.7, +0.2] \\
MATH, Pythia-6.9B & 0.93 & +13.3 & +14.0 & +13.9 & $-$0.7 [$-$1.0, $-$0.4] & $-$0.6 [$-$1.0, $-$0.3] \\
 & 0.84 & +9.2 & +9.8 & +10.0 & $-$0.7 [$-$1.0, $-$0.4] & $-$0.8 [$-$1.2, $-$0.4] \\
 & 0.67 & +4.8 & +5.6 & +5.4 & $-$0.9 [$-$1.2, $-$0.5] & $-$0.6 [$-$1.1, $-$0.1] \\
 & 0.52 & +2.4 & +3.2 & +3.2 & $-$0.8 [$-$1.1, $-$0.5] & $-$0.8 [$-$1.2, $-$0.3] \\
 & 0.40 & +0.1 & +0.8 & +0.8 & $-$0.6 [$-$1.0, $-$0.3] & $-$0.7 [$-$1.2, $-$0.2] \\
 & 0.31 & $-$1.0 & $-$0.3 & $-$0.5 & $-$0.6 [$-$1.0, $-$0.3] & $-$0.5 [$-$1.0, $-$0.0] \\
 & 0.18 & $-$1.8 & $-$1.6 & $-$1.7 & $-$0.2 [$-$0.5, +0.1] & $-$0.1 [$-$0.5, +0.3] \\
MATH, Gemma-2B & 0.74 & +3.8 & +4.2 & +2.9 & $-$0.4 [$-$0.8, $-$0.0] & +0.8 [+0.2, +1.5] \\
 & 0.66 & +2.2 & +2.2 & +0.8 & $-$0.0 [$-$0.3, +0.3] & +1.4 [+0.7, +2.1] \\
 & 0.53 & +1.0 & +1.3 & $-$0.1 & $-$0.3 [$-$0.6, +0.0] & +1.1 [+0.4, +1.8] \\
 & 0.40 & +0.3 & $-$0.1 & $-$1.2 & +0.5 [+0.1, +0.8] & +1.5 [+0.9, +2.1] \\
 & 0.32 & $-$0.8 & $-$0.5 & $-$1.1 & $-$0.3 [$-$0.7, +0.1] & +0.4 [$-$0.2, +0.9] \\
 & 0.24 & $-$0.6 & $-$0.6 & $-$1.6 & $-$0.0 [$-$0.4, +0.3] & +0.9 [+0.4, +1.5] \\
 & 0.15 & $-$0.8 & $-$0.8 & $-$1.7 & +0.0 [$-$0.3, +0.4] & +0.9 [+0.4, +1.4] \\
\bottomrule
\end{tabular}}
\end{table}

In the controlled experiment the gain declines smoothly with prior quality rather than jumping at a threshold, and for TS it crosses zero, the break-even point, at a rank correlation between 0.3 and 0.4 (Table~\ref{tab:4}); the myopic index on Gemma-2B breaks even only near 0.55. On a fixed workload the gain falls smoothly as the prior is degraded, on the Pythia workloads from +12 to +14 points at rank correlation 0.93, to +4 to +6 at 0.67, and to between +0.1 and +1.4 at about 0.4, depending on the policy. The break-even point should depend on workload hardness, the budget and the spread of pass rates, so we report it for these workloads only. The paired differences in Table~\ref{tab:4} show that no policy dominates. On the two Pythia workloads, TS is 0.3--0.9 points below UCB and the myopic index in most rows, with intervals below zero in 20 of 28 comparisons (22 counting two intervals that end at $-$0.0). On Gemma-2B, TS is 0.4--1.5 points above the myopic index, while TS and UCB are within 0.5 points of each other. The differences are an order of magnitude smaller than the effect of prior quality; this, for the three index policies we tested, is what the title summarizes.

The graded effect explains why the cheap priors fail and gives a simple rule for when to use an informed policy. Their rank correlations on the hardest workloads (0.05--0.46) lie near or below the break-even point. Proxy samples fail for an additional reason: on hard problems a small model also almost always fails, so a handful of its samples cannot separate the inputs, and their cost is not recovered. In practice, a practitioner can proceed as follows.
\begin{enumerate}
\item Build the prior from past runs on the same inputs.
\item Measure its rank correlation with pass rates on a calibration set, and expect the gain to scale with it.
\item If the correlation is low, use RR. On our workloads the break-even point for TS was 0.3--0.4, but it depends on hardness, budget and the spread of pass rates, so it should be checked on the calibration set rather than taken as a fixed threshold.
\item Use TS with a tempered prior, with the mixture weight chosen by cross-validation on the calibration set (a fixed weight of 0.75 did better in Table~\ref{tab:10}, but we chose it after seeing those results); without tempering, TS and especially UCB lost points on an easy workload (Tables~\ref{tab:9} and \ref{tab:10}), and the myopic index was the least robust on stored samples (Table~\ref{tab:2}).
\end{enumerate}

On easy workloads no policy gains, and tempering the prior limits the losses. On Weaver's MATH500, GPQA and MMLU-Pro with Llama-3.1 8B and 70B (Saad-Falcon et al., 2025), RR already solves 66--99\% of the problems. Tempering the prior by cross-validation limits the myopic index's loss to at most 1.5 points (Appendix~\ref{app:E}). Thompson Sampling, without tempering, loses at most 0.6 points on the easy Large Language Monkeys workloads (Table~\ref{tab:2}).

\textbf{On freshly generated samples from recent models, the pattern holds on hard workloads, and an untempered prior can hurt on easy ones.} To test whether replaying older stored samples matters, we generated new samples with vLLM: 64 per problem at temperature 0.7 for 200 level-4 and level-5 problems of MATH-500, graded with math-verify, from Qwen2.5-0.5B-Instruct (mean pass rate 0.16; 33\% of problems never solved), Qwen2.5-Math-1.5B-Instruct (0.62) and Qwen2.5-Math-7B-Instruct (0.74). With 0.5B as the target, the history prior gains 1.8--2.6 points at $u=4$ and the exchangeable prior none (Table~\ref{tab:9}). The gain follows the prior's rank correlation, not the size of the history model: the 7B history correlates less with the 0.5B pass rates (0.56) than the 1.5B history does (0.62) and gives the smaller gain at $u=4$. With 1.5B as the target, round-robin already solves 86--91\%, and an untempered history prior from 7B loses up to 4.2 points with TS and 10.9 with UCB: on an easy workload the remaining gain lies in the few hard inputs, and an imperfect prior withholds attempts from exactly those inputs.

\begin{table}[t]
\centering\small
\caption{Fresh samples (Qwen2.5 family; 0.5B is Qwen2.5-0.5B-Instruct, Math-1.5B and Math-7B are Qwen2.5-Math-Instruct models; hard MATH-500). Change in solved share relative to RR (points), 100 replays, bootstrap 95\% CIs.}\label{tab:9}
\resizebox{\linewidth}{!}{%
\begin{tabular}{lllllll}
\toprule
target $\leftarrow$ history & $\rho$ & u & RR solved & TS, history & UCB, history & TS, exch. \\
\midrule
0.5B $\leftarrow$ Math-1.5B & 0.62 & 4 & 36.0\% & +2.4 [+2.0, +2.8] & +2.6 [+2.2, +3.0] & $-$0.2 [$-$0.4, 0.0] \\
 & 0.62 & 16 & 58.4\% & $-$0.1 [$-$0.5, +0.3] & $-$1.8 [$-$3.1, $-$0.6] & 0.0 [$-$0.1, +0.1] \\
0.5B $\leftarrow$ Math-7B & 0.56 & 4 & 35.4\% & +1.8 [+1.4, +2.2] & +2.2 [+1.8, +2.7] & $-$0.3 [$-$0.4, $-$0.1] \\
 & 0.56 & 16 & 57.6\% & +0.3 [0.0, +0.6] & $-$0.3 [$-$1.1, +0.3] & 0.0 [$-$0.1, +0.2] \\
Math-1.5B $\leftarrow$ Math-7B & 0.77 & 4 & 86.4\% & $-$1.6 [$-$2.3, $-$1.0] & $-$5.2 [$-$6.5, $-$4.1] & 0.0 [$-$0.1, 0.0] \\
 & 0.77 & 16 & 91.0\% & $-$4.2 [$-$4.9, $-$3.4] & $-$10.9 [$-$12.1, $-$9.7] & 0.0 [0.0, 0.0] \\
\bottomrule
\end{tabular}}
\end{table}

Tempering the prior toward the exchangeable distribution removes the loss on the easy workload while keeping most of the gain on the hard one (Table~\ref{tab:10}). The pre-specified rule, choosing the weight by cross-validation on the calibration half, cut the loss on the easy workload from $-$1.7 and $-$4.1 points to $-$0.5 and 0.0, at the cost of part of the hard-workload gain (+1.6 of +2.9). A fixed weight of 0.75 on the history prior did better on both: it kept +2.4 of +2.9 points on the hard workload and avoided every loss on the easy one. We chose 0.75 after seeing these results, so it should be read as an illustration of what tempering can achieve, not as a validated default; the cross-validated weight is the rule we recommend when no such tuning is possible.

\begin{table}[t]
\centering\small
\caption{TS with a tempered history prior on fresh samples (gain over RR, points, 30 replays; no CIs). The prior of each input mixes its history-bin distribution (weight w) with the exchangeable distribution (weight 1 $-$ w). This run uses its own 30 replays, so the w = 1 column differs from the TS column of Table~\ref{tab:9} (100 replays) by up to 0.5 points. The weight 0.75 was chosen after seeing these results; the CV column is the pre-specified rule.}\label{tab:10}
\begin{tabular}{lllll}
\toprule
target $\leftarrow$ history & u & history (w = 1) & tempered (w = 0.75) & tempered (w chosen by CV) \\
\midrule
0.5B $\leftarrow$ Math-1.5B & 4 & +2.9 & +2.4 & +1.6 \\
0.5B $\leftarrow$ Math-1.5B & 16 & $-$0.3 & +0.4 & $-$0.1 \\
Math-1.5B $\leftarrow$ Math-7B & 4 & $-$1.7 & +0.4 & $-$0.5 \\
Math-1.5B $\leftarrow$ Math-7B & 16 & $-$4.1 & 0.0 & 0.0 \\
\bottomrule
\end{tabular}
\end{table}

The gain survives charging attempts by their length and missed acceptances, but not a verifier that accepts wrong answers more often than the workload is solved. We tested both relaxations of the Section~\ref{sec:2} assumptions on the four hard workloads at $u=16$ (Table~\ref{tab:5}; 20 replays each).

The first relaxation charges each attempt by its stored output length rather than counting samples, with the budget set to $u\times N$ times the mean length. The gain of TS with the history prior holds or grows, because correct samples are shorter than incorrect ones (about 300--450 against 440--650 characters), so concentrating attempts on solvable inputs also saves tokens.

The second relaxation adds verifier errors that the policies are not told about. A false rejection rate of 20\% leaves the gain within 0.2 points or larger. False acceptances are different. An accepted input leaves the pool whether or not the answer is correct, and at $u=16$ a false acceptance rate of 5\% removes most hopeless inputs from the pool with a wrong answer. Every policy then exhausts the pool, and the comparison loses its object. The gain falls by about 40\% at a false acceptance rate of 2\% and almost disappears at 5\% (at most 0.6 points; Table~\ref{tab:5}). On these workloads the mean pass rate is 0.01--0.09, so the gain requires a verifier whose false acceptance rate is well below the rate at which the workload is solved. Unit tests and proof checkers usually meet this condition; as the next experiment shows, most learned reward models on hard problems do not.

\begin{table}[t]
\centering\small
\caption{Gain of TS with the history prior over RR (points) when attempts are charged by length or the verifier errs ($u=16$, 20 replays). Columns 2--6 vary the verifier: false acceptance rate (FA) and false rejection rate (FR). The last column charges each attempt by its output length. This run uses its own replays, so the error-free column differs from Table~\ref{tab:2b} by up to 2 points.}\label{tab:5}
\resizebox{\linewidth}{!}{%
\begin{tabular}{lllllll}
\toprule
workload & no errors & FA 1\% & FA 2\% & FA 5\% & FR 20\% & length-charged \\
\midrule
MATH, Pythia-1.4B & +11.1 & +10.1 & +6.7 & +0.6 & +12.1 & +13.2 \\
MATH, Pythia-6.9B & +11.7 & +8.8 & +7.3 & +0.3 & +11.6 & +14.5 \\
MATH, Gemma-2B & +3.4 & +3.4 & +1.9 & 0.0 & +4.7 & +5.1 \\
CodeContests, Llama-3-8B & +3.3 & +2.7 & +1.9 & +0.3 & +3.1 & +3.6 \\
\bottomrule
\end{tabular}}
\end{table}

With real learned verifiers on hard problems, part of the gain survives only where both the verifier and the prior are informative (Table~\ref{tab:6}). We used the reward-model scores stored with the Weaver samples (Saad-Falcon et al., 2025) for Llama-3.1-8B on MATH500 and GPQA, restricted to problems the model solves less than 30\% of the time, with the history prior taken from Llama-3.1-70B. The acceptance threshold is set on half of the problems so that 1\% or 5\% of incorrect samples are accepted, and results are reported on all problems. We use $u=2$, because at $u=16$ round-robin with a perfect verifier already solves 90--98\% of these problems and leaves no headroom.

Most stored verifiers barely separate correct from incorrect samples on these problems. On GPQA, the nine reward models we screened have AUROC (area under the ROC curve for separating correct from incorrect samples) between 0.44 and 0.52 on problems below a 20\% pass rate (a stricter cut than the 30\% used for Table~\ref{tab:6}), and on MATH500 only the Qwen process reward model is clearly informative (AUROC 0.76; the next best, 0.60). On MATH500, that verifier keeps about half of the gain available with a perfect verifier (+1.1 to +1.7 points against +2.5), while a weaker reward model keeps less. On GPQA the history prior is itself weak on hard problems (rank correlation 0.22 with the 8B pass rates), and TS loses 2.7 points even with a perfect verifier, as Table~\ref{tab:4} predicts for a prior below the break-even point. With learned verifiers, no policy solves more than about 1\% of the GPQA problems.

\begin{table}[t]
\centering\small
\caption{Online allocation with stored reward models as the verifier ($u=2$, 10 replays). Hard subsets: problems the 8B model solves less than 30\% of the time. ``FA target'' is the share of incorrect calibration samples the threshold accepts. RR columns give the share of problems truly solved and falsely accepted; the last column is the gain of TS with the history prior over RR, with a bootstrap 95\% CI.}\label{tab:6}
\begin{tabular}{>{\raggedright\arraybackslash}p{0.267\linewidth}>{\raggedright\arraybackslash}p{0.205\linewidth}>{\raggedright\arraybackslash}p{0.065\linewidth}>{\raggedright\arraybackslash}p{0.065\linewidth}>{\raggedright\arraybackslash}p{0.065\linewidth}>{\raggedright\arraybackslash}p{0.184\linewidth}}
\toprule
workload & verifier & FA target & RR solved & RR falsely accepted & TS gain \\
\midrule
MATH500, hard (N = 173, mean p 0.13) & perfect & --- & 25.6\% & 0\% & +2.5 [+1.0, +3.8] \\
 & Qwen PRM (min) & 1\% & 3.5\% & 1.5\% & +1.1 [+0.2, +2.0] \\
 & Qwen PRM (min) & 5\% & 12.6\% & 7.2\% & +1.7 [$-$0.1, +3.5] \\
 & InternLM2 Reward 7B & 1\% & 0.5\% & 2.1\% & +0.2 [$-$0.2, +0.5] \\
 & InternLM2 Reward 7B & 5\% & 2.1\% & 8.2\% & +0.9 [+0.4, +1.4] \\
GPQA, hard (N = 414, mean p 0.12) & perfect & --- & 23.3\% & 0\% & $-$2.7 [$-$3.7, $-$1.7] \\
 & ArmorRM & 1\% & 0.5\% & 1.6\% & $-$0.1 [$-$0.3, 0.0] \\
 & ArmorRM & 5\% & 1.0\% & 7.8\% & $-$0.5 [$-$0.7, $-$0.2] \\
 & Qwen PRM (min) & 1\% & 0.3\% & 2.1\% & $-$0.1 [$-$0.2, +0.1] \\
 & Qwen PRM (min) & 5\% & 0.8\% & 7.9\% & 0.0 [$-$0.3, +0.2] \\
\bottomrule
\end{tabular}
\end{table}

\textbf{A simple model predicts when verifier errors erase the gain.} Let an attempt on input i be accepted with probability $a_i = p_i(1-\beta) + (1-p_i)\alpha$, where $\alpha$ is the false acceptance rate and $\beta$ the false rejection rate, and let an accepted input leave the pool. Two facts follow.

\emph{Priority is unchanged by verifier errors.} With known pass rates, the time to acceptance of each input is geometric with constant hazard $a_i$, and an acceptance is correct with probability $\pi_i = p_i(1-\beta)/a_i$. For such jobs the Gittins index is the reward rate $\pi_i a_i = p_i(1-\beta)$ (Gittins, 1979), so the optimal policy still attempts inputs in decreasing order of $p_i$. Verifier errors change how much there is to gain, not where to spend.

\emph{The gain shrinks to zero as false acceptances clear the pool.} Let N be the number of inputs and S = uN the total budget of attempts. In a mean-field approximation, round-robin gives every input up to k attempts, with k set by $\sum_i (1-(1-a_i)^k)/a_i = S$, and the index policy covers inputs in order of $p_i$ at an expected cost of $1/a_i$ each. Both reach the same value $\sum_i \pi_i$ once $S \ge \sum_i 1/a_i$, and a hopeless input ($p_i$ = 0) contributes 1/$\alpha$ to this sum. With $\beta$ = 0, the gain therefore disappears once $\alpha$ $\ge$ $\alpha^\ast$, where $\alpha^\ast$ solves $\frac{1}{N}\sum_i 1/(p_i + (1-p_i)\alpha) = u$; we report the threshold as $x^\ast = u\alpha^\ast$. False acceptances make round-robin abandon hopeless inputs on its own, so a workload with more hopeless inputs tolerates a larger u$\alpha$ (Table~\ref{tab:7}). The threshold $x^\ast$ is where the bound reaches zero; most of the gain is lost before it, for example at $u=16$ and a 5\% rate (u$\alpha$ = 0.8) the bound is already 0.5 points on Pythia-1.4B ($x^\ast$ = 0.88) and 0.2 on CodeContests ($x^\ast$ = 0.95). For a fixed workload the remaining gain depends on u and $\alpha$ almost only through u$\alpha$: on MATH with Pythia-1.4B, half of the error-free bound remains at u$\alpha$ $\approx$ 0.25 for $u=4$, 16 and 64 alike.

\begin{table}[t]
\centering\small
\caption{Upper bound on the gain of any policy over RR (points, known pass rates, mean-field) at $u=16$, the exhaustion threshold $x^\ast$, and the simulated TS gains of Table~\ref{tab:5}. ``Hopeless share'' is the share of inputs never solved in the stored samples. The bound collapses where the simulated gain does. MATH with Llama-3-8B is an easy workload not run in Table~\ref{tab:5}; it is listed only for its threshold.}\label{tab:7}
\begin{tabular}{>{\raggedright\arraybackslash}p{0.188\linewidth}>{\raggedright\arraybackslash}p{0.089\linewidth}>{\raggedright\arraybackslash}p{0.089\linewidth}>{\raggedright\arraybackslash}p{0.255\linewidth}>{\raggedright\arraybackslash}p{0.255\linewidth}}
\toprule
workload & hopeless share & $x^\ast = u\alpha^\ast$ & bound, FA 0 / 1\% / 2\% / 5\% & TS (Table~\ref{tab:5}), FA 0 / 1\% / 2\% / 5\% \\
\midrule
MATH, Pythia-1.4B & 0.64 & 0.88 & 17.1 / 11.6 / 7.4 / 0.5 & 11.1 / 10.1 / 6.7 / 0.6 \\
MATH, Pythia-6.9B & 0.55 & 0.84 & 17.5 / 12.0 / 7.9 / 0.2 & 11.7 / 8.8 / 7.3 / 0.3 \\
MATH, Gemma-2B & 0.34 & 0.60 & 15.6 / 9.3 / 4.7 / 0.0 & 3.4 / 3.4 / 1.9 / 0.0 \\
CodeContests, Llama-3-8B & 0.91 & 0.95 & 7.0 / 3.0 / 1.6 / 0.2 & 3.3 / 2.7 / 1.9 / 0.3 \\
MATH, Llama-3-8B & 0.15 & 0.29 & --- & --- \\
\bottomrule
\end{tabular}
\end{table}

The same calculation, given each stored reward model's measured false acceptance and true acceptance rates, predicts the share of the perfect-verifier gain that survives in Table~\ref{tab:6} (Table~\ref{tab:8}). The prediction is within 0.05 of the simulation in five of eight cases and underestimates the retained share in the other three. It also shows what limits learned verifiers on hard problems: at a 1\% false acceptance target they reject most correct answers (true acceptance 2--15\%), so the effective pass rate $p_i(1-\beta)$ is too low for any allocation to matter.

\begin{table}[t]
\centering\small
\caption{Predicted versus simulated share of the perfect-verifier gain retained with stored reward models ($u=2$). TA is the true acceptance rate, 1 $-$ $\beta$. On GPQA the perfect-verifier ``gain'' is a loss ($-$2.7 points, Table~\ref{tab:6}), so the ratio there is the share of that loss that remains.}\label{tab:8}
\resizebox{\linewidth}{!}{%
\begin{tabular}{lllllll}
\toprule
workload & verifier & FA target & measured FA & measured TA & predicted ratio & simulated TS ratio \\
\midrule
MATH500, hard & Qwen PRM & 1\% & 0.010 & 0.15 & 0.25 & 0.44 \\
MATH500, hard & Qwen PRM & 5\% & 0.050 & 0.54 & 0.67 & 0.67 \\
MATH500, hard & InternLM2 RM & 1\% & 0.012 & 0.02 & 0.03 & 0.07 \\
MATH500, hard & InternLM2 RM & 5\% & 0.050 & 0.07 & 0.12 & 0.37 \\
GPQA, hard & ArmorRM & 1\% & 0.010 & 0.03 & 0.05 & 0.05 \\
GPQA, hard & ArmorRM & 5\% & 0.050 & 0.05 & 0.09 & 0.19 \\
GPQA, hard & Qwen PRM & 1\% & 0.010 & 0.02 & 0.03 & 0.03 \\
GPQA, hard & Qwen PRM & 5\% & 0.050 & 0.03 & 0.06 & 0.02 \\
\bottomrule
\end{tabular}}
\end{table}

\section{Discussion and limitations}\label{sec:6}

We do not know why the heuristic persists, and one obvious explanation does not hold. The surveyed methods are not mainly evaluated on benchmarks their models largely solve: in an LLM-assisted reading of the 13 papers, only 3 report single-attempt accuracy of at least 50\% on most of their main benchmarks, and the median across the 9 papers that report usable numbers is about 36\%. What matters for the inverted-U, however, is the budget relative to difficulty, and the mean pass rate at the planned budget predicted the captured share on our workloads (Section~\ref{sec:2}). On those workloads the best rule captured at least half of the headroom whenever u$\cdot$$\bar{p}$ was 4 or more, and at most 13\% whenever it was 1.5 or less. Mapping the surveyed papers onto this scale, with each rule's mean budget (Appendix~\ref{app:D}, Table D1), puts the easier settings of five of the eight papers at u$\cdot$$\bar{p}$ of about 4 to 17, where ``harder gets more'' is close to optimal; these methods were probably near-optimal there. Many of their harder settings fall between 1.5 and 4, a range that none of our workloads covers, so our evidence says nothing about them. Only a few fall at or below 1.5, where the rules failed on our workloads: AIME with the smallest model in IAS (0.3), AIME25 with Qwen2.5-3B in FSC (at most 1.0), and SciQ with Phi-3-mini in one paper (1.1). The cost we measure therefore arises when such rules are carried to harder workloads than most of these papers tested: weak models, hard evaluation sets, small budgets, and RL prompt pools in which many prompts are never solved. The mapping is coarse, because a mean hides the many unsolvable inputs that make a workload hard, and several of the accuracies are self-consistency accuracies, which overstate the pass rate. The authors of IAS (Park et al., 2025), the rule we test in Section~\ref{sec:4}, note that they assign the maximum budget to the hardest problems while ``erring on the side of optimism''.

For practitioners, our results reduce to two questions: is there a verifier, and is there an informative prior? The list below condenses the step-by-step procedure of Section~\ref{sec:5} into checks that can each be run on a calibration set before any allocation rule is chosen; none of them requires new training.
\begin{enumerate}
\item With a programmatic verifier (unit tests, a proof checker or an answer key), use RR by default. If past runs on the same inputs exist, measure the prior's rank correlation with pass rates on a calibration set. On our workloads TS broke even at about 0.3--0.4, and its gain grew with the correlation (Table~\ref{tab:4}). Above that point, use the prior tempered toward the exchangeable prior. We tested tempering only with TS (Table~\ref{tab:10}), so that is the configuration we recommend, with the weight chosen by cross-validation on the calibration set; a fixed weight of 0.75 did better but was chosen after seeing the results. Untempered priors, especially with UCB, lost points on an easy workload (Table~\ref{tab:9}).
\item With a learned verifier, first estimate its false acceptance rate on hard inputs. If the budget per input times this rate approaches the threshold $x^\ast$ (Table~\ref{tab:7}), expect little or no gain from any policy.
\item Without a verifier, compute the left-branch share, or simply the mean pass rate, at the planned budget. If the workload is hard, difficulty-proportional rules will capture little of the headroom (Table~\ref{tab:1}). Use the marginal-value rule with a calibrated history signal if one exists; otherwise, do not expect any rule to beat uniform allocation by much (Section~\ref{sec:4}).
\item In every case, report results on hard workloads, not only on benchmarks the model nearly solves.
\end{enumerate}

Our conclusions rest on offline replay, an idealized verifier, four fixed policies and a partly automated survey, and each of these limits how far they extend. We treat stored samples as i.i.d. attempts, with one replication on newly generated samples (Table~\ref{tab:9}), and we model attempts as independent; if attempts on the same input are correlated, each failure carries less information than the model assumes, so the posterior updates would be overconfident. The history prior needs past runs on the same inputs; for novel inputs, the cheap priors we evaluated, including difficulty judgments by strong LLMs, gave little or no gain. The gain requires a verifier whose false acceptance rate is well below the workload's pass rate (Table~\ref{tab:5}). Most stored reward models we tested fail this condition on hard problems; a strong process reward model on MATH500 kept about half of the gain, and a weaker reward model kept a small part (Table~\ref{tab:6}). Our positive results therefore apply mainly to settings with programmatic verifiers, such as unit tests, proof checkers and answer keys. If every attempt is verified once, verification cost adds the same amount to every attempt under every policy, so it acts as a smaller budget rather than changing the comparison. ``Priors, not policies'' is established only for the four policies we tested (RR, TS, UCB and the Bayes-myopic index); a learned policy could behave differently. Two of our recommendations rest on limited evidence: the tempering weight w = 0.75 was tested on one model family with two target models (Table~\ref{tab:10}; 30 replays), and the left-branch diagnostic was checked on eleven related workload--model combinations (Section~\ref{sec:2}). We study math, code and formal proofs with models up to 70B parameters and do not establish that the same magnitudes hold for frontier reasoning models or open-ended tasks. Charging attempts by output length preserves the gain (Table~\ref{tab:5}), but we do not measure FLOPs or latency, and proxy costs are scaled by parameter count. Finally, LLM agents did the first pass of the survey coding under a fixed codebook. We re-judged that coding with full text, double-coded a subsample and checked every remaining HGM candidate condition by condition against its full text (Appendix~\ref{app:D}). These checks were LLM-assisted: an LLM proposed each verdict, we reviewed it, and we judged only the ambiguous cases manually. The confirmations are therefore not fully independent of the LLM coding. Inclusion screening of the 1,269 candidates was automated and not independently verified, so the survey may miss HGM papers, and the count of 13 is an approximate lower bound on prevalence.

Several claims were narrowed during the study. An apparent advantage of Thompson Sampling under degraded priors did not survive 100 replays with paired intervals, and we withdrew it (Table~\ref{tab:4}). The survey count fell from 19 to 13 after the condition-level full-text check (Section~\ref{sec:3}). Recomputing headroom with exact curves showed that the in-distribution headroom reported as zero by Bae (2026a) is 1.1--1.6\% (Appendix~\ref{app:B}).

Two new experiments, rather than re-analyses of existing samples, would close the main gaps. Live generation with a frontier reasoning model on a recurring benchmark, with the history prior taken from its previous release, would test whether the magnitudes transfer. Combining a history prior with a verifier that is itself calibrated on past runs, for example by requiring repeated acceptances on inputs the prior marks as hopeless, would test whether the gain can be recovered with learned verifiers. Evidence from other domains also urges caution. In a compute-value audit of video world models, no adaptive-depth policy beat uniform compute once verification was charged, although the authors note that this result is not universal (Jiang et al., 2026). Our positive results require an informative prior, so they do not contradict this finding.

\section{Related work}\label{sec:7}

Much of the work on test-time compute allocation ranks inputs by estimated difficulty and reports gains over uniform allocation; it rarely asks on which workloads, and with what information, reallocation pays off. We address that question rather than proposing a new allocator. Brown et al. (2024) document how coverage scales with repeated sampling, and Schaeffer et al. (2025) show that the per-problem failure rate under repeated sampling decays exponentially and that aggregate power laws arise from heavy-tailed difficulty. Snell et al. (2025) select a compute-optimal strategy per difficulty bin at matched budget and state that they do not account for the cost of estimating difficulty. Damani et al. (2025) learn input-adaptive allocation; their savings are 0--10\% on the full chat set and 25--40\% on a subset that combines the lowest- and highest-variance prompts. AdaBoN (Raman et al., 2025) reports that allocating in proportion to reward variance performs about as well as uniform allocation, which is consistent with our finding that dispersion must enter through marginal value. Stroebl et al. (2024) show that false positives cap what repeated sampling can gain with an imperfect verifier; our verifier-error model makes the analogous point for allocation and gives the false acceptance level at which the gain over round-robin vanishes. Iyengar (2026) fits a beta-binomial difficulty prior and a model of correlated verifier errors on a labeled pilot, combines the prior with each task's own pilot counts, and uses the resulting success formula to choose, for each budget, the generators, the number of samples and the number of verifier calls. We instead allocate attempts sequentially across inputs and show that the gain over round-robin comes from the same inputs' records under another model. Its report that per-problem solve rates of Llama-3-8B-Instruct and Gemma-7B correlate at 0.80 on MATH is independent evidence that such records are a strong signal. Its precision ceiling bounds success under false positives, whereas our threshold bounds the gain of reallocation over round-robin.

Some recent LLM methods already allocate by marginal value, and our survey codes them as non-monotone (Section~\ref{sec:3}). UAB (Nguyen et al., 2026) reallocates a fixed sampling budget across questions greedily by marginal gain on a concave coverage surrogate, using the uncertainty of initial samples, and Zhai et al. (2026) cast the allocation of a finite average budget across a batch as constrained policy optimization. Our marginal-value rule is the same greedy step applied to estimated pass rates; what we add is when such reallocation can pay and what information it needs. Fan et al. (2026) give reasoning models one token budget for a whole exam and find that they act mostly as greedy sequential solvers rather than rationing compute across questions. DORA (Wang et al., 2025) allocates rollouts among directions within one problem's search, which our survey excludes as within-input allocation.

Rollout allocation for RL with verifiable rewards is the setting closest to ours. DAPO (Yu et al., 2025) drops prompts whose rollouts are all correct or all incorrect, and online difficulty filtering (Bae et al., 2025) favors prompts with intermediate pass rates. Both act on pass rates measured on the same prompts, which is the kind of information our history prior uses. Our analysis is consistent with these choices and carries them to inference-time allocation, with a test for when such history is informative enough to use.

Our theory draws on classical results in stratified sampling, inequality measurement, optimal search and sequential index policies. Neyman (1934) derived optimal allocation for stratified sampling, Atkinson (1970) introduced the inequality index that our continuous-regime gain reduces to, Koopman (1957) and Stone (1975) developed optimal search, and Weitzman (1979) introduced index policies for sequential search. Simulation optimization contributed OCBA (Chen et al., 2000) and the knowledge gradient (Frazier et al., 2008), and work on restarts and algorithm portfolios (Luby et al., 1993; Kautz et al., 2002; Streeter et al., 2007) studied how to spend runs across instances.

Crowdsourcing reached the closest results in the binary regime. Opt-KG (Chen et al., 2013; 2015) sets highly ambiguous items aside and shows that the advantage of adaptivity depends on the budget. Khetan \& Oh (2016) show that the budget advantage of adaptive over non-adaptive schemes is $\lambda/\lambda_{\min}$, the ratio of collective to worst-case task difficulty, so the gain grows with heterogeneity.

Our survey follows the model of cross-field methodological audits such as Kapoor \& Narayanan (2023), who documented leakage across 294 papers in 17 fields.

Two earlier preprints on allocation for neural combinatorial optimization (Bae, 2026a; 2026b) motivated this study. Both concern the continuous regime, and both reported allocation gains that were difficult to disentangle from sampling noise and from the spread of difficulty. Appendix~\ref{app:B} corrects an in-distribution estimate in the former, giving the previous estimator, the source of its bias and the corrected value.

\section{Conclusion}\label{sec:8}

Giving harder inputs more attempts is the right way to divide a test-time budget only when the budget is large relative to difficulty. On public stored samples the best such rule captures half or more of the attainable gain on workloads a model mostly solves but at most 13\% on workloads it mostly fails, even with true pass rates, because the optimal static allocation is inverted-U in difficulty. With a verifier, the gain over round-robin came from information about the inputs rather than from the choice among the four policies we tested. A history prior from a related model nearly doubled the solved share on the hardest workloads at 16 attempts per problem (from 14\% to 27\% on MATH with Pythia-1.4B). The gain shrank smoothly as the prior was degraded, the cheap priors we tried gave little or no gain, and the gain required a verifier whose false acceptances are rare relative to the pass rate. The pattern held on freshly generated Qwen2.5 samples, where tempering the prior removed the losses that an untempered prior caused on an easy workload.

In practice, the choice of allocation rule should follow from the information available. Without a verifier, check the left-branch share or the mean pass rate at the planned budget before trusting a difficulty-proportional rule. With a programmatic verifier, use round-robin unless a prior from past runs clears the break-even rank correlation on a calibration set, and then use Thompson Sampling with a tempered prior. In either case, report results on hard workloads. Our evidence covers recurring inputs with binary, verifiable success in math, code and formal proofs, and does not cover frontier reasoning models or open-ended tasks. Three lines of work would extend it: (i) live generation with a frontier reasoning model on a recurring benchmark, with the prior taken from its previous release; (ii) verifiers, calibrated on past runs, that keep false acceptances rare on hard problems; and (iii) priors for novel inputs that clear the break-even rank correlation at an acceptable cost.

\section*{Data availability}

All workloads are public stored samples: Large Language Monkeys (Brown et al., 2024; released as \texttt{monkey\_business}), PRM800K (Lightman et al., 2024), Weaver (Saad-Falcon et al., 2025) and, for Appendix~\ref{app:A}, HPO-B (Pineda Arango et al., 2021). We also generated new samples from Qwen2.5 models for one replication (Table~\ref{tab:9}). Code, these samples, the survey coding with its codebooks, and the result files are available at https://github.com/nepersoned/priors-not-policies.

\paragraph{Use of generative AI tools.} Large language models were used as research tools in the survey coding and as difficulty raters; both uses are described in the main text (Sections~3 and~5). LLM-based assistants were also used to help write analysis code and to edit the text. The author reviewed all outputs and takes full responsibility for the content of the article.

\clearpage
\appendix
\section{Continuous objectives}\label{app:A}

\subsection{Model}

For best-of-k on a continuous score, write the expected loss after k attempts as
f\_i(k) = $a_i$ + b\_i$\cdot$k\^{}($-$$\alpha$).

\textbf{Closed form.} Assume a common $\alpha$ and ignore integrality. Optimal allocation gives $k_i$ $\propto$ b\_i\^{}(1/(1+$\alpha$)), and the reducible gain of optimal over uniform allocation is

\[ 1 - M_p(b)/M_1(b),\qquad p = 1/(1+\alpha), \]

where M\_p is the power mean. This is the Atkinson (1970) inequality index of b with aversion $\varepsilon=\alpha/(1+\alpha)$. It does not depend on the budget.

\textbf{Consequences.}
\begin{itemize}
\item A common rescaling of all b\_i (a change of difficulty \emph{level}) leaves the index unchanged.
\item Level can matter only through the irreducible share $a_i$, or through $\alpha$ varying with difficulty.
\item Neyman allocation is the case $\alpha$ = 1 (variance $\propto$ 1/n, optimal n $\propto$ standard deviation).
\item If $\log b\sim\mathcal N(\mu,\sigma^2)$, the gain is $1-\exp(-\varepsilon\sigma^2/2)$; $\mu$ does not appear.
\end{itemize}

\subsection{Fit and prediction (exact curves from stored pools)}

\begin{table}[t]
\centering\small

\begin{tabular}{>{\raggedright\arraybackslash}p{0.706\linewidth}>{\raggedright\arraybackslash}p{0.065\linewidth}>{\raggedright\arraybackslash}p{0.065\linewidth}>{\raggedright\arraybackslash}p{0.065\linewidth}}
\toprule
domain & r(true headroom, Atkinson closed form) & r(true, optimum on fitted curves) & mean true headroom (\%) \\
\midrule
NCO cluster AM & 0.97 & 0.99 & 9.83 \\
NCO cluster SymNCO & 0.97 & 1.00 & 7.15 \\
NCO size AM & 0.92 & 0.97 & 0.83 \\
NCO size SymNCO & 0.93 & 0.98 & 0.77 \\
HPO-B all pairs & 0.80 & 0.93 & 13.80 \\
LLM GSM8K 4o-mini RM internlm & 0.96 & 0.98 & 1.27 \\
LLM GSM8K 4o-mini RM skyworks & 0.97 & 0.98 & 0.87 \\
LLM MATH PRM800K ORM & 0.77 & 0.98 & 19.36 \\
LLM MATH PRM800K PRM & 0.78 & 0.96 & 14.62 \\
\bottomrule
\end{tabular}
\end{table}

The fitted-curve optimum tracks the true headroom closely. The closed form, which assumes a single common $\alpha$, is weaker where $\alpha$ varies across inputs (HPO-B, PRM800K).

\subsection{Spread vs level across 13 domain $\times$ objective cells}

\textbf{Method.} Windowed sets of inputs are formed, and the true headroom is computed per set. The table reports partial Spearman correlations of headroom with spread (the SD of log b; for binary objectives, of log rate) and with level (median difficulty), each controlling for the other. Values are medians over 10 window seeds, with the IQR in brackets.

\begin{table}[t]
\centering\small

\resizebox{\linewidth}{!}{%
\begin{tabular}{llllll}
\toprule
cell & objective & power-law R$^2$ & exp. R$^2$ & spread given level & level given spread \\
\midrule
NCO cluster AM & continuous & 0.996 & 0.490 & +0.89 [+0.89, +0.91] & +0.41 [+0.26, +0.49] \\
NCO cluster SymNCO & continuous & 0.996 & 0.521 & +0.91 [+0.90, +0.91] & +0.34 [+0.24, +0.39] \\
NCO size AM & continuous & 0.998 & 0.568 & +0.54 [+0.51, +0.60] & -0.71 [-0.73, -0.64] \\
NCO size SymNCO & continuous & 0.999 & 0.568 & +0.80 [+0.74, +0.81] & -0.58 [-0.70, -0.51] \\
HPO-B all pairs & continuous & 0.992 & 0.723 & +0.76 [+0.75, +0.77] & +0.59 [+0.58, +0.66] \\
LLM GSM8K 4o-mini RM internlm & continuous & 0.998 & 0.784 & +0.93 [+0.93, +0.94] & +0.24 [+0.14, +0.28] \\
LLM GSM8K 4o-mini RM skyworks & continuous & 0.999 & 0.797 & +0.50 [+0.46, +0.53] & -0.26 [-0.27, -0.17] \\
LLM MATH PRM800K ORM & continuous & 0.991 & 0.634 & +0.50 [+0.47, +0.53] & -0.82 [-0.86, -0.76] \\
LLM MATH PRM800K PRM & continuous & 0.976 & 0.695 & +0.76 [+0.72, +0.77] & -0.49 [-0.53, -0.40] \\
LLM MATH PRM800K binary & binary & 0.979 & 1.000 & -0.44 [-0.49, -0.32] & +0.88 [+0.86, +0.89] \\
LLM GSM8K Llama8B binary & binary & 0.986 & 1.000 & -0.34 [-0.49, -0.20] & +0.76 [+0.70, +0.81] \\
LLM MATH Llama8B binary & binary & 0.935 & 1.000 & +0.34 [+0.26, +0.43] & +0.90 [+0.89, +0.90] \\
LLM MATH Llama70B binary & binary & 0.979 & 1.000 & -0.75 [-0.76, -0.72] & +0.92 [+0.91, +0.92] \\
\bottomrule
\end{tabular}}
\end{table}

\begin{figure}[t]
\centering
\includegraphics[width=0.85\linewidth]{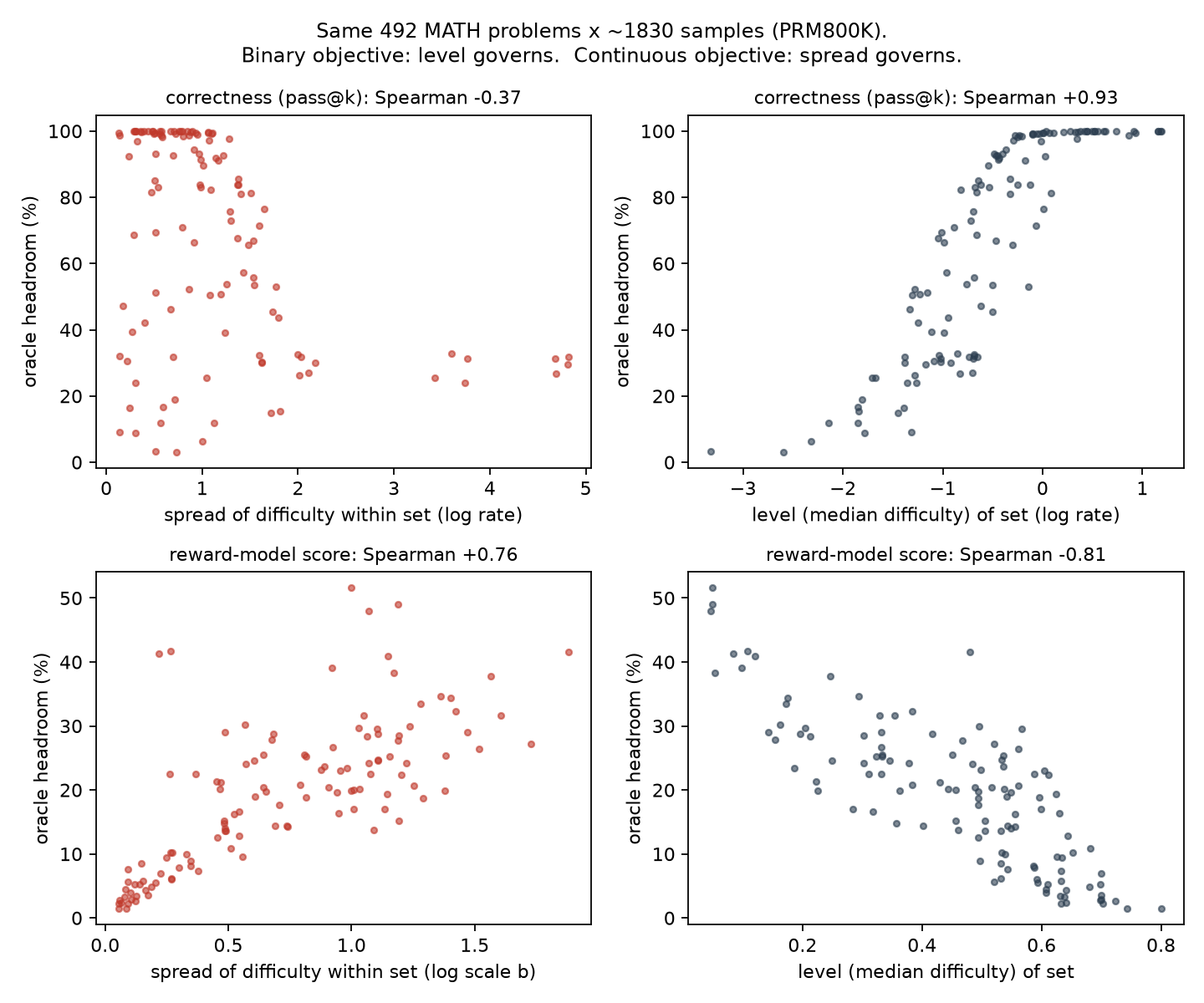}
\caption{Same LLM samples, two objectives: with correctness (binary) the level of difficulty governs the value of reallocation, and with a reward-model score (continuous) its spread does.}\label{fig:A1}
\end{figure}

\textbf{Reading.}
\begin{itemize}
\item In every binary cell, level carries the signal.
\item In every continuous cell, spread has a positive partial correlation with headroom (+0.50 to +0.93).
\item Level's partial correlation in continuous cells is sign-unstable across domains ($-$0.82 to +0.59), and in two cells (NCO size AM, PRM800K ORM) its magnitude exceeds spread's. Spread therefore has a consistent direction, but it does not always dominate; level enters through curve shape (a\_i and $\alpha$), which the closed form holds fixed.
\item On the same LLM samples (PRM800K, Weaver), switching from correctness to a reward-model score flips which statistic governs (Figure~\ref{fig:A1}).
\end{itemize}

\section{Measurement notes and a correction to Bae (2026a)}\label{app:B}

\textbf{Exact curves.} For a stored pool of n samples, the expected best-of-k is computed exactly over all orderings (a U-statistic), not by averaging a few hundred random permutations. Monte Carlo curves add noise that a greedy allocator exploits, which inflates in-sample headroom.

\textbf{Correction to Bae (2026a)}: uniform TSP-100, N = 50, S/N = 100; Axis B, a deterministic pool of 800, so exact curves give the true headroom.

\begin{table}[t]
\centering\small

\begin{tabular}{>{\raggedright\arraybackslash}p{0.171\linewidth}>{\raggedright\arraybackslash}p{0.171\linewidth}>{\raggedright\arraybackslash}p{0.192\linewidth}>{\raggedright\arraybackslash}p{0.171\linewidth}>{\raggedright\arraybackslash}p{0.171\linewidth}}
\toprule
solver & reported in-sample estimate (MC curves) & true headroom (exact curves) & sample-split estimate & mean max MC curve error (pts) \\
\midrule
POMO & 2.227 & \textbf{1.568} & 0.493 & 0.46 \\
AM & 2.567 & \textbf{1.122} & 0.168 & 3.68 \\
SymNCO & 2.206 & \textbf{1.193} & 0.215 & 6.93 \\
\bottomrule
\end{tabular}
\end{table}

\textbf{What went wrong.}
\begin{itemize}
\item The earlier in-sample estimate overstated the headroom by 0.7--1.4 points because of Monte Carlo curve noise.
\item The sample-split estimate understated it, because half-pool curves are not full-pool curves.
\item The earlier conclusion of no detectable in-distribution headroom is therefore wrong: the true value is 1.1--1.6\%.
\item Recomputed with exact curves, the out-of-distribution result stands: 14.1\% (AM) and 12.7\% (SymNCO) with in-sample exact curves, and 13.9\% and 12.0\% split.
\end{itemize}

\textbf{Lesson used in this paper.} All headroom numbers in Sections~\ref{sec:4}--\ref{sec:5} use exact curves from stored pools; sample splitting is not used to estimate headroom.

\section{Proof of the inverted-U optimum}\label{app:C}

\paragraph{Setup.} Input $i$ has pass rate $p_i \in (0,1)$ and rate $r_i = -\ln(1-p_i) > 0$, so the failure probability after $k$ attempts is $(1-p_i)^k = e^{-r_i k}$. We maximize $\sum_i \bigl[1 - e^{-r_i k_i}\bigr]$ subject to $\sum_i k_i = S$ and $k_i \ge 0$, relaxing $k_i$ to be continuous; the integer problem is solved exactly by greedy allocation on marginal values, because the objective is separable and concave.

\paragraph{Lemma (marginal value).} The marginal value of one more attempt is $\Delta(p,k) = p(1-p)^k$, and
\[
\frac{\partial \Delta}{\partial p} = (1-p)^{k-1}\bigl(1-(k+1)p\bigr).
\]
Hence $\Delta$ increases on $[0, 1/(k+1)]$ and decreases on $[1/(k+1), 1]$, with its maximum at $p^\ast = 1/(k+1)$.

\paragraph{Proposition.} The optimal allocation is
\[
k^\ast(r) = \max\Bigl(0,\ \frac{\ln(r/\lambda)}{r}\Bigr),
\]
where $\lambda > 0$ is fixed by the budget. It has the following properties.
\begin{enumerate}
\item[(a)] $k^\ast(r) = 0$ for $r \le \lambda$: these inputs are not worth attempting at this budget.
\item[(b)] Since $\frac{d}{dr}\bigl[\ln(r/\lambda)/r\bigr] = \bigl(1-\ln(r/\lambda)\bigr)/r^2$, $k^\ast$ increases with $r$ (easier inputs receive more) for $\lambda < r < e\lambda$, and decreases with $r$ (harder inputs receive more) for $r > e\lambda$.
\item[(c)] The maximum is at $r = e\lambda$, where $k^\ast = 1/(e\lambda)$.
\item[(d)] $\lambda$ decreases as the budget grows, so the maximum moves toward harder inputs as the budget grows.
\end{enumerate}

\paragraph{Minimum number of attempts.} Our experiments give every input at least one attempt. Replacing $k_i \ge 0$ by $k_i \ge 1$ changes only the boundary: wherever $k_i > 1$ the condition is again $r_i e^{-r_i k_i} = \lambda$, so
\[
k^\ast(r) = \max\Bigl(1,\ \frac{\ln(r/\lambda)}{r}\Bigr),
\]
with $\lambda$ refitted so that $\sum_i k^\ast(r_i) = S$. An input receives attempts beyond the first if and only if $\ln(r/\lambda)/r > 1$, that is, $r e^{-r} > \lambda$. Since $r e^{-r}$ is maximal at $r = 1$, this set is an interval of $r$: both hopeless inputs (small $r$) and nearly certain ones (large $r$) stop at the minimum. Where $k^\ast > 1$ the allocation is the same function $\ln(r/\lambda)/r$ as above, so (b) and (c) are unchanged and the peak remains at $r = e\lambda$. The diagnostic below uses the unconstrained form at budget $S = uN$.

\paragraph{Corollary.} A rule in which $k$ decreases with $p$ moves budget in the optimal direction only on the right branch, $r > e\lambda$. When most inputs have $r_i < e\lambda$, which at uniform allocation means that $u\, r_i$ is small (a hard workload or a small budget), every such rule moves budget in the wrong direction.

\paragraph{Proof.} The objective is separable and strictly concave in each $k_i$. The Lagrangian conditions give $r_i e^{-r_i k_i} = \lambda$ wherever $k_i > 0$; solving for $k_i$ gives the formula, and complementary slackness gives $k_i = 0$ when $r_i \le \lambda$. Differentiating in $r$ gives (b) and (c). The sum $\sum_i k^\ast(r_i;\lambda)$ is continuous and strictly decreasing in $\lambda$, so $\lambda$ is unique for each $S>0$ and decreases as $S$ grows, which gives (d). $\blacksquare$

\paragraph{Diagnostic.} Given estimated rates $\hat r_i$ on a calibration set, compute $\lambda$ at the planned budget and report the share of inputs on the left branch, $\#\{i : \hat r_i < e\lambda\}/N$. A large share means that a difficulty-proportional rule moves budget in the wrong direction for most inputs; a share near zero means that such a rule is close to optimal. Section~\ref{sec:2} reports how this share relates to the captured headroom on our workloads.

\section{Papers allocating repeated attempts by ``harder gets more'' (HGM)}\label{app:D}

\emph{HGM is labelled E2 in the codebook and the survey files; the two names refer to the same strict definition.}

\begin{figure}[t]
\centering
\includegraphics[width=0.85\linewidth]{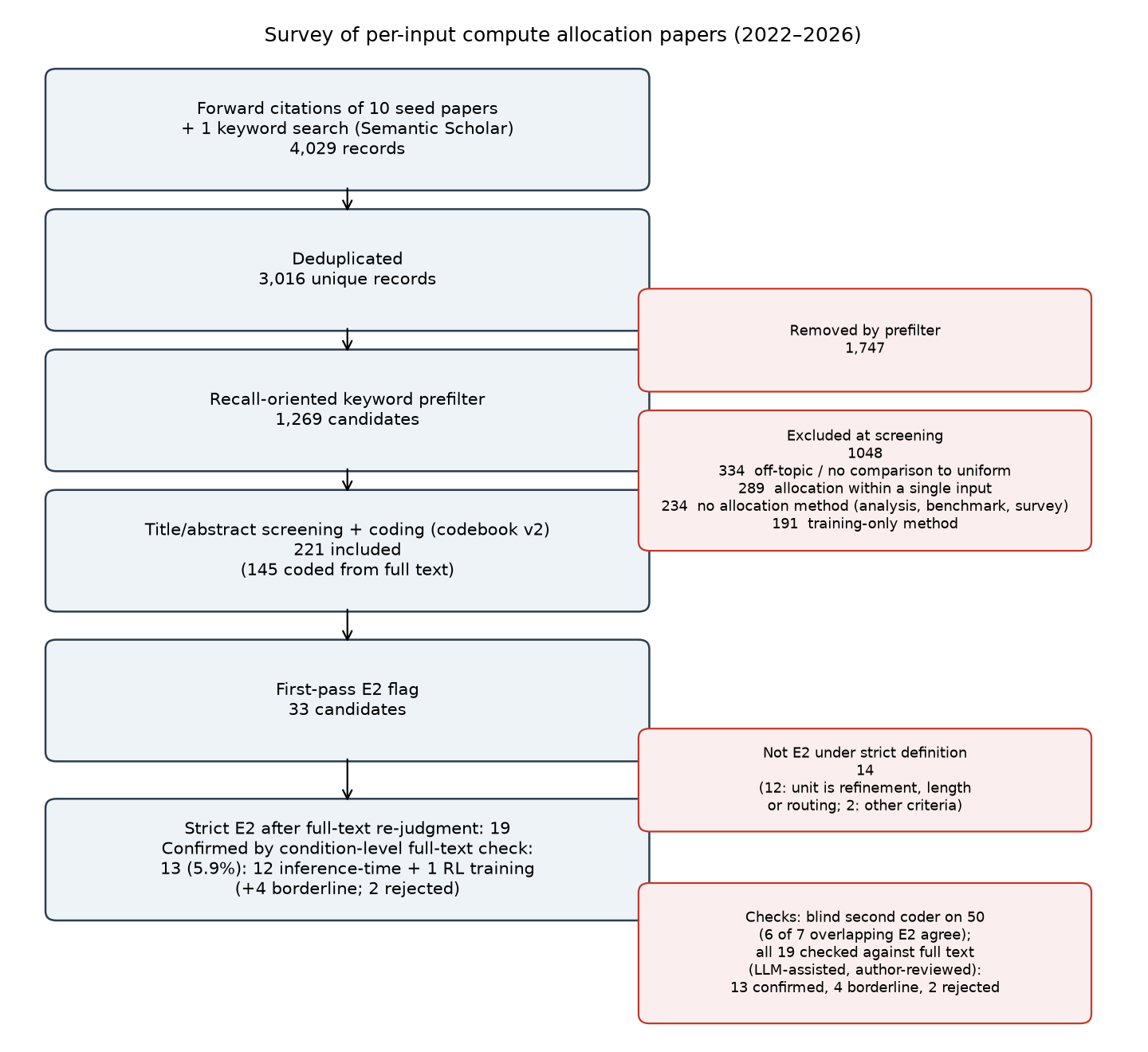}
\caption{Survey flow. Screening-exclusion categories are grouped from coders' free-text reasons by keyword, so they are approximate.}\label{fig:5}
\end{figure}

\textbf{Strict definition.} A paper qualifies only if all four hold:
\begin{itemize}
\item (a) the allocated unit is repeated independent attempts on the same input;
\item (b) a difficulty-level signal gives harder inputs more attempts, monotonically;
\item (c) there is no discount, abstention or deprioritization for hopeless inputs (a global cap reachable by every input is not a discount);
\item (d) evaluation is against uniform or fixed allocation.
\end{itemize}

\textbf{Process.}
\begin{itemize}
\item 33 candidates were flagged by first-pass coding of 221 included papers. All were re-judged with full text (32 of 33 opened).
\item A blind second coder recoded a stratified random sample of 50 included papers (15 first-pass E2 candidates, 35 others). Agreement on the E2 flag was 41 of 50 (Cohen's $\kappa$ = 0.51). The second coder flagged 8 as E2. Of the 7 that were also first-pass candidates, the strict re-judgment agrees on 6.
\item \textbf{Full-text check of all 19.} All 19 papers kept by the re-judgment were checked against their full text, condition by condition. The check was LLM-assisted: an LLM read each paper and proposed a verdict with section-level evidence, the authors reviewed every verdict, and the authors judged the ambiguous cases themselves against the full text. 
\begin{itemize}
\item In the random sample of 10, all 7 strict-E2 papers were confirmed. Of the 3 disagreements, two concern whether refinement rounds count as repeated attempts, and the third (aborting runs predicted to fail) was marked unclear.
\item Across all 19, the check confirmed 13, left 4 borderline and rejected 2.
\end{itemize}
\item Fragments below are verbatim from the source ($\le$15 words) unless marked \emph{paraphrase}.
\end{itemize}

\textbf{Counting.}
\begin{itemize}
\item \textbf{Confirmed (13, 5.9\% of 221).} All four conditions hold.
\begin{itemize}
\item 12 allocate at inference time, each with an independent rule.
\item 1 allocates RL training rollouts (marked \dag{}).
\end{itemize}
\item \textbf{Borderline (4).} One condition is unclear.
\begin{itemize}
\item Including them gives 17 (7.7\%).
\item The 4 are CoBa, AttnRL\dag{}, ODAR and Distributional PRM\ddag{}. \ddag{} marks a reuse of the IAS rule, so it is not an independent rule.
\end{itemize}
\item \textbf{Rejected by the full-text check (2).} Dynasor and Gated GeoBoN.
\end{itemize}

\begin{table}[t]
\centering\small

\begin{tabular}{>{\raggedright\arraybackslash}p{0.065\linewidth}>{\raggedright\arraybackslash}p{0.173\linewidth}>{\raggedright\arraybackslash}p{0.108\linewidth}>{\raggedright\arraybackslash}p{0.176\linewidth}>{\raggedright\arraybackslash}p{0.176\linewidth}>{\raggedright\arraybackslash}p{0.151\linewidth}}
\toprule
\# & check & paper (arXiv) & allocation rule & verbatim fragment & section \\
\midrule
1 & confirmed & Make Every Penny Count (DSC), 2408.13457 & Easy: 1 sample. Hard: resample until a stopping criterion, else cap 40. & ``For problems belonging to easy part, a single CoT sampling is performed'' & Section~\ref{sec:2}, Section~\ref{sec:2}.3 \\
2 & confirmed & Evolving in-context demonstrations, 2604.21018 & Unsolved queries keep sampling each round until T. Elimination uses a ground-truth oracle. & ``focuses compute on harder queries as easier ones are resolved'' & Section~\ref{sec:4}.2.2, Alg. 1 \\
3 & confirmed & Think Right, Not More, 2509.22101 & Level-0 claims: one shot. Level-1 claims (including persistently misclassified ones): expanded best-of-N. & ``one-shot inference for level-0 claims, and expanded reasoning for level-1'' & Section~\ref{sec:3}.2 \\
4 & borderline (a) & CoBa, 2608.07424 & One extra candidate at a time until the top answer is stable and highly scored, up to max. The extra compute also routes the top-K candidates to a strong verifier. & ``Generate one additional candidate and score it'' & Alg. 1 (Section~\ref{sec:4}) \\
5 & confirmed & Scaling with Confidence (CAS), 2607.01612 & Sample until stop-probability exceeds a threshold or reaches N. The threshold is tuned on evaluation sets. & ``otherwise, the inference continues with next generation'' & Section~\ref{sec:3}.3, Alg. 1 \\
6\dag{} & borderline (a) & AttnRL, 2509.26628 & Trees per problem = $\mathrm{Round}(\exp(-\text{pass rate})\times 6)$; zero pass rate gets the maximum. RL training. The unit is tree branches from shared prefixes. & ``we expand different number of trees according to problem difficulty'' & Section~\ref{sec:3}.2.1, Eq. 9 \\
7 & borderline (a) & ODAR, 2602.23681 & Hard: 1 fast + 5 slow best-of-N; easy: 1 fast. Attempts are confounded with routing to a stronger model. & ``Hard uses c=6 (one Fast generation plus N=5 Slow candidates'' & Section~\ref{sec:3}.1 \\
8 & confirmed & LiLaVe, 2504.16760 & Low verifier score on the probe triggers n extra samples for majority vote. & ``generate n additional samples to perform a majority voting'' & Section~\ref{sec:3}.2 \\
9 & confirmed & Know What You Don't Know (IAS), 2506.09338 & $N=\lceil\log(1-C)/\log(1-p)\rceil$, capped at $N_{\max}$. & ``challenging problems receive larger budgets up to $N_{\max}$'' & Section~\ref{sec:4}.1 \\
10 & confirmed & Self-Calibration, 2503.00031 & Sample one by one until a response meets the confidence threshold. & ``until a response meets the confidence threshold'' & Section~\ref{sec:4}.1 \\
11 & \textbf{rejected (c)} & Certaindex / Dynasor, 2412.20993 & Low certaindex runs to the resource cap, but out-of-reach or unpromising queries are terminated early, which is a discount for hopeless inputs. & ``or reaches the maximum resource cap, the scheduler denies further'' & App. D.2, Section~\ref{sec:1} \\
12 & confirmed & Text2Cypher adaptive inference, 2609.02324 & Easy/medium/hard tiers get 21/30/45 traces. & ``hard questions receive the full budget of the baseline'' & Section~\ref{sec:3}.1, Section~\ref{sec:4}.2.2 \\
13 & confirmed & FSC, 2608.24590 & $N_{\mathrm{adj}}=\lceil 1+(N-1)\hat H/\log_2 N\rceil$, monotone in predicted entropy. & ``up to N reasoning paths to difficult questions with high entropy'' & Section~\ref{sec:3}.2, Eq. 4 \\
14 & confirmed & Interpretable adaptive sampling, 2608.03961 & Fuzzy rules. High complexity or low confidence raises N, with a floor N $\ge$ min(6, $N_{\max}$). & ``High complexity or low confidence pushes the output toward larger budgets.'' & Section~\ref{sec:3}.2--3.3 \\
15\ddag{} & borderline (d) & Distributional PRM, 2605.06785 & Reuses the IAS rule of \#9. No fixed-budget comparison found. & \emph{paraphrase:} allocates budget from estimated success probability, capped at $N_{\max}$ & Section~\ref{sec:3} \\
16 & confirmed & LLM Already Knows (difficulty-aware repeated sampling), 2509.12886 & Easy: 1 direct answer. Hard: K samples for self-consistency or best-of-N. & ``easy questions are addressed with single sampling'' & Section~\ref{sec:4} \\
17 & confirmed & ACTSC, 2602.09438 & Hard items sample in windows until confidence $\ge\gamma$. The threshold $\tau$ is set per evaluation dataset. & ``until $\mathrm{Conf}(a^\ast)\ge\gamma$'' (Alg. 2) & Section~\ref{sec:3}.2, Alg. 2 \\
18\dag{} & confirmed & DATPO, 2609.08650 & $K=\lceil K_{\max}(1-V)\rceil$, $B=\lceil B_{\max}(1-V)\rceil$. Unsolved problems get the maximum. RL training. & ``concentrates exploration on unsolved problems to maximize reasoning coverage'' & Section~\ref{sec:3}.1, Eq. 1--2 \\
19 & \textbf{rejected (a)} & Gated GeoBoN, 2607.17454 & Best-of-N over candidate actions at decision points where the rollout looks inconsistent: step-level search within an episode, not repeated attempts. & ``invokes GeoBoN only when the initial rollout appears internally inconsistent'' & Abstract, Section~\ref{sec:4}.3 \\
\bottomrule
\end{tabular}
\end{table}

\textbf{Budget relative to difficulty in the surveyed papers.} Table D1 places the main settings of the confirmed papers on the u$\cdot$$\bar{p}$ scale of Section~\ref{sec:2}. Here u is the mean number of attempts per input that the paper's own adaptive rule used, and $\bar{p}$ is the mean single-attempt accuracy. All numbers were checked against the papers' PDFs. For reference, on our workloads at $u=16$ the best rule captured at least half of the headroom when u$\cdot$$\bar{p}$ $\ge$ 4 and at most 13\% when u$\cdot$$\bar{p}$ $\le$ 1.5 (Table~\ref{tab:1}); no workload of ours lies between 1.5 and 4. Where a paper reports only self-consistency accuracy (SC@N), which is at least the single-attempt accuracy, u$\cdot$$\bar{p}$ is an upper bound. Where it reports cost or tokens rather than samples, u is approximated by the rule's share of the uniform cost times N. Four papers report no usable accuracy, and one reports macro-F1; they are omitted.

\emph{Table D1. Budget relative to difficulty (u$\cdot$$\bar{p}$) in the main settings of the surveyed HGM papers. u is the adaptive rule's mean budget per input.}

\begin{table}[t]
\centering\small

\begin{tabular}{>{\raggedright\arraybackslash}p{0.194\linewidth}>{\raggedright\arraybackslash}p{0.227\linewidth}>{\raggedright\arraybackslash}p{0.227\linewidth}>{\raggedright\arraybackslash}p{0.152\linewidth}>{\raggedright\arraybackslash}p{0.076\linewidth}}
\toprule
paper & setting & u (mean, per input) & $\bar{p}$ (measure) & u$\cdot$$\bar{p}$ \\
\midrule
Self-Calibration (Huang et al., 2025) & 3 tasks $\times$ 3 models & 16 (matched) & 0.58--0.90 (pass@1) & 9--14 \\
IAS (Park et al., 2025) & MATH500, 6 models & 15--41 & 0.23--0.80 (pass@1) & 9--15 \\
IAS (Park et al., 2025) & AIME24--25, 6 models & 58--64 & 0.004--0.14 (pass@1) & 0.3--8.7 \\
DSC, 2408.13457 & 6 tasks, GPT-4 / GPT-3.5 & $\approx$ 6--33 (from cost) & 0.49--1.00 (SC@40) & $\le$ 6--17 \\
CAS, 2607.01612 & in-domain, OOD text, OOD multimodal & 5.2--10.9 & 0.45--0.71 (pass@1) & 3.2--7.0 \\
FSC (Yeom et al., 2026) & MATH500, AMC23, 4 models & $\approx$ 9--22 (from tokens) & 0.50--0.82 (SC@40) & $\le$ 7--13 \\
FSC (Yeom et al., 2026) & AIME24--25, GPQA-D, 4 models & $\approx$ 10--31 (from tokens) & 0.03--0.46 (SC@40) & $\le$ 1.0--9.6 \\
ACTSC, 2602.09438 & MATH500, GPQA-D, MMLU-Pro & 5.2--9.7 & 0.37--0.68 (SC@40) & $\le$ 3.2--6.0 \\
ACTSC, 2602.09438 & AIME24--25 & 18--30 & 0.07--0.17 (SC@40) & $\le$ 2.0--4.4 \\
Interpretable adaptive sampling, 2608.03961 & GSM8K, MATH, SciQ, 2 models & 6.9--7.9 & 0.15--0.60 (pass@1) & 1.1--4.7 \\
DATPO, 2609.08650 (RL) & MATH training set, 2 base models & $\le$ 16 rollouts & 0.26--0.40 (base avg@8 on MATH500) & $\le$ 4.2--6.4 at the start of training \\
\bottomrule
\end{tabular}
\end{table}

\textbf{Notes.}
\begin{itemize}
\item Most of these papers report gains over uniform allocation, but not mainly on benchmarks their models largely solve: in an LLM-assisted reading, only 3 of the 13 report single-attempt accuracy of at least 50\% on most main benchmarks (median about 36\% over the 9 with usable numbers). Section 6 discusses why gains can still appear at large budgets.
\item Paper \#9 explicitly acknowledges the choice: \emph{``assign the maximum budget to the hardest problems''}, describing this as ``erring on the side of optimism'' (Section~\ref{sec:4} intro).
\item Correction to the coding file: \#19's title ends ``via Zero-Shot Geometric Verification''.
\end{itemize}

\section{Full online results and robustness}\label{app:E}

\subsection{All 18 workload $\times$ budget cells (20 replays; change in solved share vs RR, points)}

\begin{table}[t]
\centering\small

\resizebox{\linewidth}{!}{%
\begin{tabular}{llllllll}
\toprule
workload & u & RR solved & TS exch. & UCB exch. & TS history & UCB history & myopic history \\
\midrule
MATH\_Pythia-1.4B & 4 & 4.8\% & -0.4 & +0.0 & \textbf{+7.9} & +8.0 & +7.9 \\
MATH\_Pythia-1.4B & 16 & 14.1\% & -0.2 & +0.0 & \textbf{+12.5} & +12.7 & +12.1 \\
MATH\_Pythia-1.4B & 64 & 34.7\% & -0.1 & +0.0 & \textbf{+12.3} & +12.2 & +11.8 \\
MATH\_Pythia-6.9B & 4 & 6.5\% & -0.5 & +0.0 & \textbf{+8.7} & +9.3 & +8.9 \\
MATH\_Pythia-6.9B & 16 & 18.6\% & +0.3 & +0.0 & \textbf{+13.8} & +14.1 & +13.6 \\
MATH\_Pythia-6.9B & 64 & 41.6\% & -0.1 & +0.0 & \textbf{+12.3} & +13.5 & +13.9 \\
MATH\_Gemma-2B & 4 & 23.4\% & +0.1 & +0.0 & \textbf{+6.3} & +6.2 & +5.5 \\
MATH\_Gemma-2B & 16 & 48.2\% & +0.1 & +0.0 & \textbf{+2.7} & +3.4 & +2.6 \\
MATH\_Gemma-2B & 64 & 72.4\% & -0.1 & +0.0 & \textbf{+2.1} & +0.7 & -2.7 \\
CodeContests\_Llama-3-8B-Instruct & 4 & 3.3\% & +0.0 & +0.0 & \textbf{+2.5} & +2.5 & +2.6 \\
CodeContests\_Llama-3-8B-Instruct & 16 & 5.5\% & -0.1 & +0.0 & \textbf{+4.0} & +3.8 & +3.7 \\
CodeContests\_Llama-3-8B-Instruct & 64 & 8.6\% & +0.1 & +0.0 & \textbf{+3.8} & +3.8 & +3.9 \\
MiniF2F-MATH\_Llama-3-8B-Instruct & 4 & 32.9\% & -0.4 & +0.0 & \textbf{+4.1} & +5.5 & +4.5 \\
MiniF2F-MATH\_Llama-3-8B-Instruct & 16 & 40.0\% & +0.2 & +0.0 & \textbf{+2.1} & +1.7 & +0.2 \\
MiniF2F-MATH\_Llama-3-8B-Instruct & 64 & 43.3\% & +0.0 & +0.0 & \textbf{+0.1} & -0.2 & -1.0 \\
MATH\_Llama-3-8B-Instruct & 4 & 53.4\% & +0.0 & +0.0 & \textbf{+6.4} & +8.4 & +8.4 \\
MATH\_Llama-3-8B-Instruct & 16 & 78.2\% & +0.1 & +0.0 & \textbf{+1.1} & +0.9 & -0.3 \\
MATH\_Llama-3-8B-Instruct & 64 & 90.3\% & +0.0 & +0.0 & \textbf{-0.6} & -1.0 & -2.0 \\
\bottomrule
\end{tabular}}
\end{table}

\subsection{Cross-validated tempering of the prior (Bayes-myopic)}

The weight w on the history prior, versus an exchangeable pool, is chosen by inner cross-validation on the calibration half.

\begin{table}[t]
\centering\small

\begin{tabular}{>{\raggedright\arraybackslash}p{0.436\linewidth}>{\raggedright\arraybackslash}p{0.065\linewidth}>{\raggedright\arraybackslash}p{0.065\linewidth}>{\raggedright\arraybackslash}p{0.065\linewidth}>{\raggedright\arraybackslash}p{0.065\linewidth}>{\raggedright\arraybackslash}p{0.065\linewidth}>{\raggedright\arraybackslash}p{0.065\linewidth}}
\toprule
workload pair & u & RR solved & history & CV-tempered & chosen w & E2 online \\
\midrule
MATH\_Pythia-1.4B <- MATH\_Pythia-6.9B & 4 & 4.8\% & +8.5 & +7.5 & 0.61 & +0.0 \\
MATH\_Pythia-1.4B <- MATH\_Pythia-6.9B & 16 & 14.7\% & +11.5 & +10.8 & 0.76 & +0.0 \\
MATH\_Pythia-1.4B <- MATH\_Pythia-6.9B & 64 & 33.4\% & +13.2 & +11.9 & 0.86 & -0.3 \\
MATH\_Gemma-2B <- MATH\_Llama-3-8B-Instruct & 4 & 25.7\% & +4.2 & +3.8 & 0.70 & +0.0 \\
MATH\_Gemma-2B <- MATH\_Llama-3-8B-Instruct & 16 & 46.2\% & +2.9 & +1.8 & 0.56 & -0.6 \\
MATH\_Gemma-2B <- MATH\_Llama-3-8B-Instruct & 64 & 71.1\% & -3.2 & -1.0 & 0.46 & -1.6 \\
MATH\_Llama-3-8B-Instruct <- MATH\_Llama-3-70B-Instruct & 4 & 54.4\% & +8.7 & +7.2 & 0.84 & +0.0 \\
MATH\_Llama-3-8B-Instruct <- MATH\_Llama-3-70B-Instruct & 16 & 76.8\% & +0.2 & +0.1 & 0.53 & -1.1 \\
MATH\_Llama-3-8B-Instruct <- MATH\_Llama-3-70B-Instruct & 64 & 89.8\% & -2.4 & -0.2 & 0.38 & -0.4 \\
MATH\_Llama-3-8B-Instruct <- MATH\_Pythia-1.4B & 4 & 54.4\% & +1.2 & +1.3 & 0.49 & +0.0 \\
MATH\_Llama-3-8B-Instruct <- MATH\_Pythia-1.4B & 16 & 77.1\% & -1.1 & -0.3 & 0.45 & -0.4 \\
MATH\_Llama-3-8B-Instruct <- MATH\_Pythia-1.4B & 64 & 90.6\% & -4.1 & -0.9 & 0.38 & +0.0 \\
MATH\_Llama-3-70B-Instruct <- MATH\_Llama-3-8B-Instruct & 4 & 76.7\% & +1.4 & +0.7 & 0.57 & -0.3 \\
MATH\_Llama-3-70B-Instruct <- MATH\_Llama-3-8B-Instruct & 16 & 93.4\% & -2.9 & -0.7 & 0.36 & -0.3 \\
MATH\_Llama-3-70B-Instruct <- MATH\_Llama-3-8B-Instruct & 64 & 97.2\% & -0.4 & -0.5 & 0.29 & +0.0 \\
CodeContests\_Llama-3-8B-Instruct <- CodeContests\_Llama-3-70B-Instruct & 4 & 3.5\% & +2.1 & +1.7 & 0.68 & +0.0 \\
CodeContests\_Llama-3-8B-Instruct <- CodeContests\_Llama-3-70B-Instruct & 16 & 5.8\% & +3.9 & +3.3 & 0.74 & +0.0 \\
CodeContests\_Llama-3-8B-Instruct <- CodeContests\_Llama-3-70B-Instruct & 64 & 8.6\% & +3.9 & +3.6 & 0.82 & -1.5 \\
CodeContests\_Llama-3-70B-Instruct <- CodeContests\_Llama-3-8B-Instruct & 4 & 11.4\% & +2.0 & +1.0 & 0.61 & +0.0 \\
CodeContests\_Llama-3-70B-Instruct <- CodeContests\_Llama-3-8B-Instruct & 16 & 16.6\% & +0.7 & +0.3 & 0.45 & -1.1 \\
CodeContests\_Llama-3-70B-Instruct <- CodeContests\_Llama-3-8B-Instruct & 64 & 21.3\% & -0.5 & -0.4 & 0.42 & -2.4 \\
MiniF2F-MATH\_Llama-3-8B-Instruct <- MiniF2F-MATH\_Llama-3-70B-Instruct & 4 & 32.2\% & +5.2 & +5.1 & 0.76 & +0.0 \\
MiniF2F-MATH\_Llama-3-8B-Instruct <- MiniF2F-MATH\_Llama-3-70B-Instruct & 16 & 39.7\% & +0.6 & +0.9 & 0.63 & -4.4 \\
MiniF2F-MATH\_Llama-3-8B-Instruct <- MiniF2F-MATH\_Llama-3-70B-Instruct & 64 & 43.0\% & -0.4 & +0.0 & 0.28 & -3.9 \\
MiniF2F-MATH\_Llama-3-70B-Instruct <- MiniF2F-MATH\_Llama-3-8B-Instruct & 4 & 30.8\% & +4.9 & +3.9 & 0.78 & +0.0 \\
MiniF2F-MATH\_Llama-3-70B-Instruct <- MiniF2F-MATH\_Llama-3-8B-Instruct & 16 & 39.2\% & -0.9 & -1.7 & 0.61 & -4.0 \\
MiniF2F-MATH\_Llama-3-70B-Instruct <- MiniF2F-MATH\_Llama-3-8B-Instruct & 64 & 43.1\% & -1.1 & -0.3 & 0.44 & -5.0 \\
GSM8K\_Llama-3-8B-Instruct <- GSM8K\_Llama-3-70B-Instruct & 4 & 99.8\% & -0.3 & -0.3 & 0.10 & +0.0 \\
GSM8K\_Llama-3-8B-Instruct <- GSM8K\_Llama-3-70B-Instruct & 16 & 100.0\% & +0.0 & +0.0 & 0.03 & +0.0 \\
GSM8K\_Llama-3-8B-Instruct <- GSM8K\_Llama-3-70B-Instruct & 64 & 100.0\% & +0.0 & +0.0 & 0.00 & +0.0 \\
GSM8K\_Llama-3-70B-Instruct <- GSM8K\_Llama-3-8B-Instruct & 4 & 99.6\% & -0.7 & -0.5 & 0.08 & +0.0 \\
GSM8K\_Llama-3-70B-Instruct <- GSM8K\_Llama-3-8B-Instruct & 16 & 100.0\% & +0.0 & +0.0 & 0.08 & +0.0 \\
GSM8K\_Llama-3-70B-Instruct <- GSM8K\_Llama-3-8B-Instruct & 64 & 100.0\% & +0.0 & +0.0 & 0.05 & +0.0 \\
\bottomrule
\end{tabular}
\end{table}

\subsection{Easy workloads (Weaver)}

\begin{table}[t]
\centering\small

\resizebox{\linewidth}{!}{%
\begin{tabular}{llllll}
\toprule
workload pair & u & RR solved & history & CV-tempered & chosen w \\
\midrule
MATH500 8B <- 70B & 4 & 93.5\% & -0.0 & +0.0 & 0.52 \\
MATH500 8B <- 70B & 16 & 99.2\% & -1.4 & -0.0 & 0.00 \\
MATH500 8B <- 70B & 64 & 99.2\% & -0.9 & -0.0 & 0.00 \\
MATH500 70B <- 8B & 4 & 97.9\% & -2.0 & -0.3 & 0.24 \\
MATH500 70B <- 8B & 16 & 98.6\% & -3.2 & +0.0 & 0.00 \\
MATH500 70B <- 8B & 64 & 98.6\% & -2.3 & +0.0 & 0.00 \\
GPQA 8B <- 70B & 4 & 68.5\% & +0.4 & +0.4 & 0.49 \\
GPQA 8B <- 70B & 16 & 95.1\% & -4.9 & +0.0 & 0.01 \\
GPQA 8B <- 70B & 64 & 95.1\% & -2.5 & +0.0 & 0.00 \\
GPQA 70B <- 8B & 4 & 65.8\% & -0.0 & +0.0 & 0.42 \\
GPQA 70B <- 8B & 16 & 77.5\% & +0.1 & +0.1 & 0.47 \\
GPQA 70B <- 8B & 64 & 80.7\% & +0.0 & +0.0 & 0.00 \\
MMLU-Pro 8B <- 70B & 4 & 83.2\% & -2.2 & -1.5 & 0.50 \\
MMLU-Pro 8B <- 70B & 16 & 96.8\% & -6.7 & +0.0 & 0.00 \\
MMLU-Pro 8B <- 70B & 64 & 96.8\% & -2.3 & +0.0 & 0.00 \\
MMLU-Pro 70B <- 8B & 4 & 87.4\% & -0.1 & -0.0 & 0.57 \\
MMLU-Pro 70B <- 8B & 16 & 92.0\% & -1.6 & +0.0 & 0.00 \\
MMLU-Pro 70B <- 8B & 64 & 92.0\% & -1.9 & +0.0 & 0.00 \\
\bottomrule
\end{tabular}}
\end{table}

\subsection{Sensitivity of the history prior to the number of bins (TS-history, $u=16$, 30 replays)}

\begin{table}[t]
\centering\small

\begin{tabular}{llll}
\toprule
workload & 3 bins & 5 bins & 10 bins \\
\midrule
MATH\_Pythia-1.4B & +11.0 & +12.4 & +11.6 \\
MATH\_Pythia-6.9B & +12.2 & +13.1 & +13.4 \\
MATH\_Gemma-2B & +3.0 & +5.1 & +2.2 \\
CodeContests\_Llama-3-8B-Instruct & +2.1 & +3.7 & +3.9 \\
\bottomrule
\end{tabular}
\end{table}

\subsection{Charged proxy: number of proxy samples (TS, $u=16$, 30 replays)}

\begin{table}[t]
\centering\small

\begin{tabular}{lllll}
\toprule
target $\leftarrow$ proxy & m = 4 & m = 8 & m = 16 & m = 32 \\
\midrule
MATH\_Pythia-1.4B & -1.0 & -1.9 & -3.7 & -7.9 \\
MATH\_Pythia-6.9B & -1.2 & -1.8 & -2.6 & -3.4 \\
\bottomrule
\end{tabular}
\end{table}

Every proxy size loses to RR, and larger probes lose more because their cost grows faster than their information.

\subsection{Left-branch diagnostic}

\begin{itemize}
\item n = 11 workload--model combinations.
\item Spearman -0.92, bootstrap 95\% CI [-0.97, -0.58].
\item Pearson -0.96.
\item Leave-one-out range -0.95 to -0.89.
\end{itemize}

\subsection{Imperfect verifier, full grid ($u=16$, 20 replays)}

FA is the false acceptance rate and FR the false rejection rate. ``RR solved'' counts only truly correct acceptances. ``falsely accepted'' is the share of inputs that left the pool with a wrong answer under RR.

\begin{table}[t]
\centering\small

\begin{tabular}{>{\raggedright\arraybackslash}p{0.346\linewidth}>{\raggedright\arraybackslash}p{0.065\linewidth}>{\raggedright\arraybackslash}p{0.065\linewidth}>{\raggedright\arraybackslash}p{0.065\linewidth}>{\raggedright\arraybackslash}p{0.065\linewidth}>{\raggedright\arraybackslash}p{0.065\linewidth}>{\raggedright\arraybackslash}p{0.065\linewidth}>{\raggedright\arraybackslash}p{0.065\linewidth}}
\toprule
workload & FR & FA & RR solved & RR falsely accepted & TS exch. $-$ RR & TS history $-$ RR & TS history falsely accepted \\
\midrule
MATH\_Pythia-1.4B & 0\% & 0\% & 14.6\% & 0.0\% & -0.3 & +11.1 & 0.0\% \\
MATH\_Pythia-1.4B & 0\% & 1\% & 13.6\% & 15.4\% & +0.2 & +10.1 & 17.3\% \\
MATH\_Pythia-1.4B & 0\% & 2\% & 14.2\% & 30.3\% & -0.1 & +6.7 & 31.3\% \\
MATH\_Pythia-1.4B & 0\% & 5\% & 12.7\% & 79.4\% & +0.1 & +0.6 & 78.4\% \\
MATH\_Pythia-1.4B & 0\% & 10\% & 8.4\% & 91.6\% & +0.0 & +0.0 & 91.6\% \\
MATH\_Pythia-1.4B & 20\% & 0\% & 11.2\% & 0.0\% & -0.2 & +12.1 & 0.0\% \\
MATH\_Pythia-1.4B & 20\% & 1\% & 11.8\% & 16.9\% & -0.2 & +8.7 & 15.7\% \\
MATH\_Pythia-1.4B & 20\% & 2\% & 11.5\% & 30.6\% & -0.1 & +6.5 & 30.9\% \\
MATH\_Pythia-1.4B & 20\% & 5\% & 10.6\% & 77.8\% & -0.0 & +0.6 & 77.8\% \\
MATH\_Pythia-1.4B & 20\% & 10\% & 6.2\% & 93.8\% & +0.0 & +0.0 & 93.8\% \\
MATH\_Pythia-6.9B & 0\% & 0\% & 19.6\% & 0.0\% & -0.0 & +11.7 & 0.0\% \\
MATH\_Pythia-6.9B & 0\% & 1\% & 18.5\% & 16.6\% & -0.1 & +8.8 & 16.1\% \\
MATH\_Pythia-6.9B & 0\% & 2\% & 17.7\% & 30.2\% & -0.1 & +7.3 & 30.7\% \\
MATH\_Pythia-6.9B & 0\% & 5\% & 17.4\% & 78.0\% & -0.1 & +0.3 & 77.9\% \\
MATH\_Pythia-6.9B & 0\% & 10\% & 11.8\% & 88.2\% & +0.0 & +0.0 & 88.2\% \\
MATH\_Pythia-6.9B & 20\% & 0\% & 17.8\% & 0.0\% & +0.1 & +11.6 & 0.0\% \\
MATH\_Pythia-6.9B & 20\% & 1\% & 16.8\% & 15.9\% & -0.3 & +9.0 & 16.4\% \\
MATH\_Pythia-6.9B & 20\% & 2\% & 16.8\% & 30.5\% & -0.1 & +6.4 & 31.0\% \\
MATH\_Pythia-6.9B & 20\% & 5\% & 15.3\% & 76.8\% & -0.1 & +0.6 & 76.9\% \\
MATH\_Pythia-6.9B & 20\% & 10\% & 10.1\% & 89.9\% & +0.0 & +0.0 & 89.9\% \\
MATH\_Gemma-2B & 0\% & 0\% & 45.5\% & 0.0\% & +0.0 & +3.4 & 0.0\% \\
MATH\_Gemma-2B & 0\% & 1\% & 44.6\% & 14.6\% & +0.2 & +3.4 & 15.1\% \\
MATH\_Gemma-2B & 0\% & 2\% & 44.7\% & 30.9\% & +0.0 & +1.9 & 30.3\% \\
MATH\_Gemma-2B & 0\% & 5\% & 38.2\% & 61.8\% & +0.0 & +0.0 & 61.8\% \\
MATH\_Gemma-2B & 0\% & 10\% & 30.2\% & 69.8\% & +0.0 & +0.0 & 69.8\% \\
MATH\_Gemma-2B & 20\% & 0\% & 41.4\% & 0.0\% & +0.0 & +4.7 & 0.0\% \\
MATH\_Gemma-2B & 20\% & 1\% & 41.6\% & 14.4\% & +0.3 & +3.4 & 14.9\% \\
MATH\_Gemma-2B & 20\% & 2\% & 40.9\% & 30.2\% & -0.0 & +1.8 & 30.5\% \\
MATH\_Gemma-2B & 20\% & 5\% & 35.9\% & 63.9\% & +0.0 & +0.0 & 63.7\% \\
MATH\_Gemma-2B & 20\% & 10\% & 27.4\% & 72.6\% & +0.0 & +0.0 & 72.6\% \\
CodeContests\_Llama-3-8B-Instruct & 0\% & 0\% & 5.6\% & 0.0\% & +0.0 & +3.3 & 0.0\% \\
CodeContests\_Llama-3-8B-Instruct & 0\% & 1\% & 5.8\% & 16.6\% & +0.2 & +2.7 & 16.0\% \\
CodeContests\_Llama-3-8B-Instruct & 0\% & 2\% & 5.2\% & 30.8\% & +0.1 & +1.9 & 30.1\% \\
CodeContests\_Llama-3-8B-Instruct & 0\% & 5\% & 5.1\% & 78.5\% & +0.0 & +0.3 & 78.4\% \\
CodeContests\_Llama-3-8B-Instruct & 0\% & 10\% & 4.4\% & 95.6\% & +0.0 & +0.0 & 95.6\% \\
CodeContests\_Llama-3-8B-Instruct & 20\% & 0\% & 5.1\% & 0.0\% & -0.1 & +3.1 & 0.0\% \\
CodeContests\_Llama-3-8B-Instruct & 20\% & 1\% & 4.9\% & 16.9\% & +0.1 & +2.5 & 16.3\% \\
CodeContests\_Llama-3-8B-Instruct & 20\% & 2\% & 4.8\% & 31.4\% & +0.0 & +1.4 & 31.6\% \\
CodeContests\_Llama-3-8B-Instruct & 20\% & 5\% & 4.9\% & 78.9\% & -0.0 & +0.1 & 79.8\% \\
CodeContests\_Llama-3-8B-Instruct & 20\% & 10\% & 3.8\% & 96.2\% & +0.0 & +0.0 & 96.2\% \\
\bottomrule
\end{tabular}
\end{table}

\subsection{Length-charged cost ($u=16$ budget in characters, 20 replays)}

\begin{table}[t]
\centering\small

\begin{tabular}{>{\raggedright\arraybackslash}p{0.382\linewidth}>{\raggedright\arraybackslash}p{0.065\linewidth}>{\raggedright\arraybackslash}p{0.065\linewidth}>{\raggedright\arraybackslash}p{0.065\linewidth}>{\raggedright\arraybackslash}p{0.065\linewidth}>{\raggedright\arraybackslash}p{0.065\linewidth}>{\raggedright\arraybackslash}p{0.119\linewidth}}
\toprule
workload & mean length (chars) & correct samples & incorrect samples & Spearman(length, p) & RR solved & TS history $-$ RR (sd) \\
\midrule
MATH\_Pythia-1.4B & 606 & 349 & 609 & -0.27 & 14.4\% & +13.2 (2.5) \\
MATH\_Pythia-6.9B & 644 & 454 & 647 & -0.18 & 20.2\% & +14.5 (3.0) \\
MATH\_Gemma-2B & 551 & 323 & 574 & -0.47 & 44.8\% & +5.1 (2.4) \\
CodeContests\_Llama-3-8B-Instruct & 440 & 299 & 443 & -0.17 & 6.0\% & +3.6 (1.4) \\
\bottomrule
\end{tabular}
\end{table}

Correct samples are shorter than incorrect ones in every workload, and inputs with lower pass rates produce longer outputs. Concentrating attempts on solvable inputs therefore also saves tokens.

\subsection{Stored reward models as verifiers: discrimination (Llama-3.1-8B samples)}

AUROC is computed over all samples, and over problems with an 8B pass rate below 0.2 (``hard''). FA and TA are the false and true acceptance rates on held-out problems when the threshold targets $\alpha$ of incorrect calibration samples (20 random splits).

\begin{table}[t]
\centering\small

\resizebox{\linewidth}{!}{%
\begin{tabular}{llllllll}
\toprule
task & reward model & AUROC & AUROC (hard) & FA at $\alpha$=1\% & TA at $\alpha$=1\% & FA at $\alpha$=5\% & TA at $\alpha$=5\% \\
\midrule
GPQA & weaver & 0.51 & 0.50 & 2.7\% & 2.7\% & 12.4\% & 13.1\% \\
GPQA & Skyworks & 0.57 & 0.52 & 2.3\% & 4.7\% & 4.8\% & 9.9\% \\
GPQA & ArmorRM & 0.57 & 0.52 & 1.0\% & 5.1\% & 5.4\% & 11.7\% \\
GPQA & QRM & 0.58 & 0.51 & 2.5\% & 4.8\% & 5.2\% & 10.0\% \\
GPQA & GRM & 0.58 & 0.52 & 2.3\% & 4.0\% & 5.2\% & 8.9\% \\
GPQA & URM & 0.57 & 0.50 & 2.2\% & 4.4\% & 4.8\% & 10.0\% \\
GPQA & QwenPRM\_min & 0.59 & 0.44 & 0.9\% & 6.2\% & 5.1\% & 12.3\% \\
GPQA & EurusPRMStage2\_min & 0.55 & 0.50 & 9.4\% & 11.0\% & 9.3\% & 11.0\% \\
GPQA & InternLM2Reward7B & 0.60 & 0.51 & 1.0\% & 3.9\% & 5.0\% & 8.3\% \\
MATH500 & Skyworks & 0.50 & 0.48 & 2.2\% & 2.9\% & 5.1\% & 6.7\% \\
MATH500 & ArmorRM & 0.45 & 0.49 & 1.8\% & 3.5\% & 5.0\% & 8.3\% \\
MATH500 & QRM & 0.51 & 0.48 & 2.2\% & 3.0\% & 5.0\% & 7.2\% \\
MATH500 & GRM & 0.64 & 0.60 & 2.2\% & 2.8\% & 5.0\% & 6.6\% \\
MATH500 & URM & 0.43 & 0.51 & 1.6\% & 3.4\% & 4.9\% & 8.8\% \\
MATH500 & QwenPRM\_min & 0.88 & 0.76 & 1.0\% & 36.8\% & 5.1\% & 71.7\% \\
MATH500 & EurusPRMStage2\_min & 0.67 & 0.59 & 1.0\% & 14.6\% & 5.2\% & 31.0\% \\
MATH500 & InternLM2Reward7B & 0.63 & 0.50 & 1.1\% & 4.1\% & 5.0\% & 13.3\% \\
\bottomrule
\end{tabular}}
\end{table}

\subsection{Stored reward models as verifiers at $u=16$}

At $u=16$, round-robin with a perfect verifier already solves 90--98\% of the hard subsets, so no policy has room to gain. Table~\ref{tab:6} in the main text therefore uses $u=2$.

\begin{table}[t]
\centering\small

\begin{tabular}{lllll}
\toprule
task & verifier & RR solved & RR falsely accepted & TS history $-$ RR [95\% CI] \\
\midrule
MATH500 & perfect & 97.7\% & 0.0\% & -0.3 [-0.8, +0.0] \\
MATH500 & QwenPRM\_min, FA target 0.01 & 18.2\% & 9.7\% & +0.0 [-0.3, +0.3] \\
MATH500 & QwenPRM\_min, FA target 0.05 & 46.0\% & 38.4\% & -0.1 [-0.6, +0.5] \\
MATH500 & InternLM2Reward7B, FA target 0.01 & 3.7\% & 16.2\% & +0.2 [-0.1, +0.6] \\
MATH500 & InternLM2Reward7B, FA target 0.05 & 11.7\% & 53.0\% & +0.2 [-0.2, +0.6] \\
GPQA & perfect & 90.0\% & 0.0\% & +0.0 [-0.2, +0.2] \\
GPQA & ArmorRM, FA target 0.01 & 3.2\% & 9.9\% & +0.1 [-0.1, +0.2] \\
GPQA & ArmorRM, FA target 0.05 & 4.3\% & 27.6\% & +0.1 [+0.0, +0.3] \\
GPQA & QwenPRM\_min, FA target 0.01 & 2.1\% & 11.3\% & +0.1 [+0.0, +0.2] \\
GPQA & QwenPRM\_min, FA target 0.05 & 4.2\% & 30.0\% & +0.0 [-0.1, +0.1] \\
\bottomrule
\end{tabular}
\end{table}

\subsection{Captured share of the oracle gain}

\begin{figure}[t]
\centering
\includegraphics[width=0.85\linewidth]{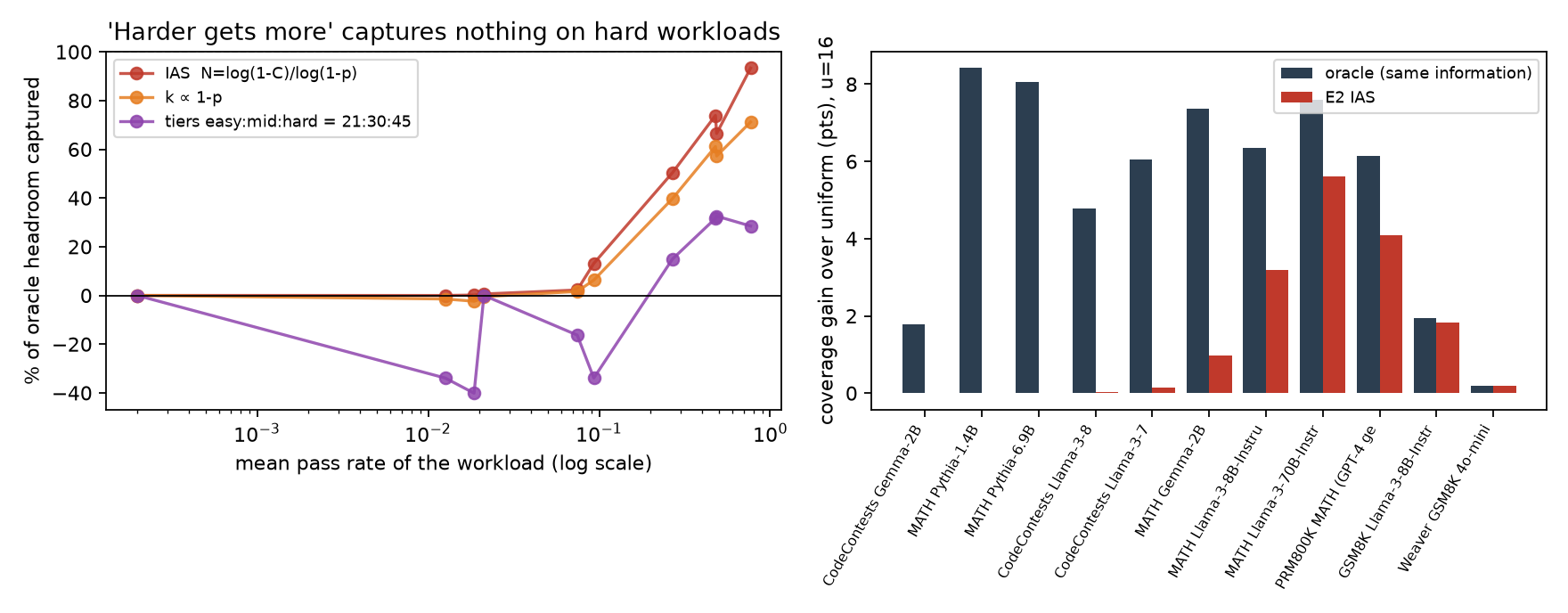}
\caption{Left: share of oracle headroom captured by ``harder gets more'' rules versus workload pass rate. Right: oracle and IAS coverage gains at $u=16$.}\label{fig:2}
\end{figure}

\subsection{Uncertainty of the main online comparison}

\begin{table}[t]
\centering\small
\caption{Uncertainty at $u=16$: 100 replays, mean change in solved share vs RR, with bootstrap 95\% CIs.}\label{tab:2b}
\begin{tabular}{>{\raggedright\arraybackslash}p{0.166\linewidth}>{\raggedright\arraybackslash}p{0.149\linewidth}>{\raggedright\arraybackslash}p{0.210\linewidth}>{\raggedright\arraybackslash}p{0.175\linewidth}>{\raggedright\arraybackslash}p{0.175\linewidth}}
\toprule
workload & TS, exchangeable & \textbf{TS, history} & UCB, history & myopic, history \\
\midrule
MATH, Pythia-1.4B & $-$0.1 [$-$0.3, 0.0] & \textbf{+12.0 [+11.5, +12.5]} & +12.0 [+11.4, +12.5] & +12.0 [+11.4, +12.5] \\
MATH, Pythia-6.9B & $-$0.1 [$-$0.2, +0.1] & \textbf{+13.7 [+13.1, +14.3]} & +14.2 [+13.6, +14.8] & +13.9 [+13.2, +14.6] \\
MATH, Gemma-2B & $-$0.1 [$-$0.3, +0.1] & \textbf{+3.4 [+2.9, +4.0]} & +3.7 [+3.1, +4.3] & +2.6 [+1.8, +3.4] \\
CodeContests, Llama-3-8B & 0.0 [$-$0.1, +0.1] & \textbf{+3.5 [+3.3, +3.8]} & +3.5 [+3.3, +3.8] & +3.5 [+3.2, +3.7] \\
MiniF2F, Llama-3-8B & 0.0 [$-$0.1, 0.0] & \textbf{+1.9 [+1.6, +2.2]} & +1.4 [+1.0, +1.7] & +0.2 [$-$0.5, +0.7] \\
MATH, Llama-3-8B & +0.1 [0.0, +0.2] & \textbf{+0.6 [+0.2, +0.9]} & +0.3 [$-$0.1, +0.7] & $-$0.2 [$-$0.6, +0.3] \\
\bottomrule
\end{tabular}
\end{table}

\subsection{Gain versus prior quality across prior sources}

\begin{figure}[t]
\centering
\includegraphics[width=0.85\linewidth]{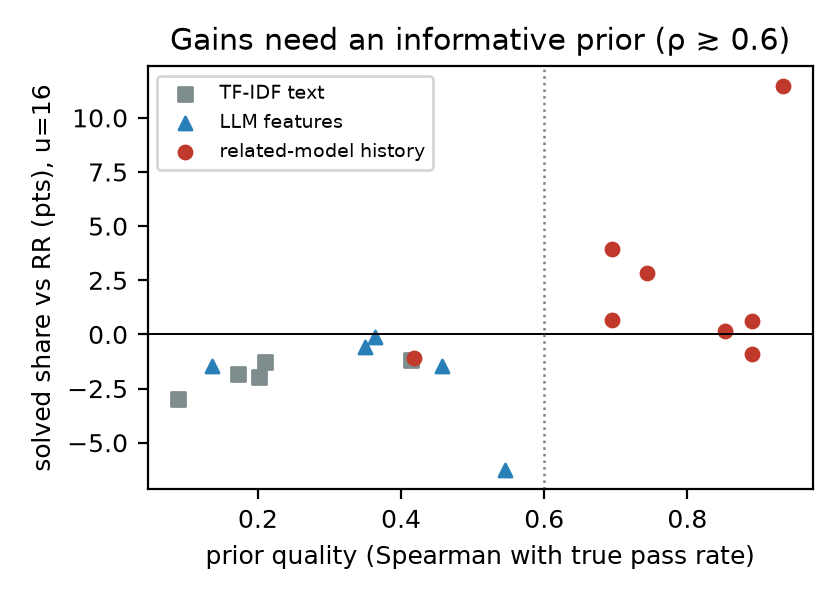}
\caption{Gain versus prior quality at $u=16$, across prior sources and workloads (observational; Bayes-myopic policy, whereas the history-prior gains in Table~\ref{tab:3} are for TS). The controlled experiment in Table~\ref{tab:4} isolates the effect of prior quality on a fixed workload.}\label{fig:4}
\end{figure}

\end{document}